\documentclass[10pt,twocolumn,twoside]{IEEEtran}
\usepackage{graphics,multirow,amsmath,amssymb,textcomp,subfigure,multirow,xspace,arydshln,cite}
\usepackage[pdftex]{graphicx}
\usepackage{booktabs}
\usepackage{caption}
\DeclareGraphicsExtensions{.png,.pdf,.jpeg,.png}
\usepackage{enumerate}
\usepackage{amsmath, amsfonts}
\usepackage{layout}
\usepackage{color}
\usepackage[ruled,vlined,boxed,linesnumbered]{algorithm2e}
\usepackage[pagebackref=true,breaklinks=true,letterpaper=true,colorlinks,bookmarks=false,linkcolor=black,citecolor=black,urlcolor=blue]{hyperref}
\LinesNumberedHidden
\DontPrintSemicolon
\makeatletter
\DeclareRobustCommand\onedot{\futurelet\@let@token\@onedot}
\def\@onedot{\ifx\@let@token.\else.\null\fi\xspace}

\makeatother

\begin{document}

\title{NIMA: Neural Image Assessment}
\author{Hossein Talebi and Peyman Milanfar\\
\thanks{H. Talebi and P. Milanfar are with Google Research, Mountain View, USA, Email: \{htalebi, milanfar\}@google.com.}
}

\maketitle
\vspace{-0cm}

\begin{abstract}
\vspace{-0 mm}
Automatically learned quality assessment for images has recently become a hot topic due to its usefulness in a wide variety of applications such as evaluating image capture pipelines, storage techniques and sharing media. Despite the subjective nature of this problem, most existing methods only predict the \textit{mean} opinion score provided by datasets such as AVA~\cite{murray2012ava} and TID2013~\cite{ponomarenko2013color}. Our approach differs from others in that we predict the \textit{distribution} of human opinion scores using a convolutional neural network. Our architecture also has the advantage of being significantly simpler than other methods with comparable performance. Our proposed approach relies on the success (and retraining) of proven, state-of-the-art deep object recognition networks. Our resulting network can be used to not only score images reliably and with high correlation to human perception, but also to assist with adaptation and optimization of photo editing/enhancement algorithms in a photographic pipeline. All this is done without need for a ``golden'' reference image, consequently allowing for single-image, semantic- and perceptually-aware, \textit{no-reference} quality assessment. 
\end{abstract}


\section{Introduction}
\label{sec:Introduction}
\vspace{0 mm}

Quantification of image quality and aesthetics have been a long-standing problem in image processing and computer vision. While technical quality assessment deals with measuring low-level degradations such as noise, blur, compression artifacts, etc., aesthetic assessment quantifies semantic level characteristics associated with emotions and beauty in images. In general, image quality assessment can be categorized into full-reference and no-reference approaches. While availability of a reference image is assumed in the former (metrics such as PSNR, SSIM\cite{wang2004image}, etc.), typically blind (no-reference) approaches rely on a statistical model of distortions to predict image quality. The main goal of both categories is to predict a quality score that correlates well with human perception. Yet, the subjective nature of image quality remains the fundamental issue. Recently, more complex models such as deep convolutional neural networks (CNNs) have been used to address this problem\cite{xue2013learning, kang2014convolutional, bosse2016deep, bianco2016use, lu2015deep, kao2015visual, mai2016composition, jin2016image}. Emergence of labeled data from human ratings has encouraged these efforts~\cite{murray2012ava,ponomarenko2013color,sheikh2005live,larson2010most,kong2016photo}. In a typical deep CNN approach, weights are initialized by training on classification related datasets (e.g.\ ImageNet~\cite{krizhevsky2012imagenet}), and then fine tuned on annotated data for perceptual quality assessment tasks.

 \subsection{Related Work}
Machine learning has shown promising success in predicting technical quality of images~\cite{xue2013learning, kang2014convolutional, bosse2016deep, bianco2016use}. Kang et. al.~\cite{kang2014convolutional} show that extracting high level features using CNNs can result in state-of-the-art blind quality assessment performance. It appears that replacing hand-crafted features with an end-to-end feature learning system is the main advantage of using CNNs for pixel-level quality assessment tasks~\cite{kang2014convolutional, bosse2016deep}. The proposed method in~\cite{kang2014convolutional} is a shallow network with one convolutional layer and two fully-connected layers, and input patches are of size $32\times32$. Bosse et al.~\cite{bosse2016deep} use a deep CNN with 12 layers to improve on image quality predictions of~\cite{kang2014convolutional}. Given the small input size ($32\times32$ patch), both methods require score aggregation across the whole image. Bianco et al. in~\cite{bianco2016use} propose a deep quality predictor based on AlexNet~\cite{krizhevsky2012imagenet}. Multiple CNN features are extracted from image crops of size $227\times227$, and then regressed to the human scores.

Success of CNNs on object recognition tasks has significantly benefited the research on aesthetic assessment. This seems natural, as semantic level qualities are directly related to image content. Recent CNN-based methods \cite{lu2015deep, kao2015visual, jin2016image, mai2016composition, kim2017deep} show a significant performance improvement compared to earlier works based on hand-crafted features~\cite{murray2012ava}. Murray et al.~\cite{murray2012ava} is the benchmark on aesthetic assessment. They introduce the AVA dataset and propose a technique to use manually designed features for style classification. Later, Lu et al.~\cite{lu2015deep, lu2015rating} show that deep CNNs are well suited to the aesthetic assessment task. Their double-column CNN~\cite{lu2015rating} consists of four convolutional and two fully-connected layers, and its inputs are the resized image and cropped windows of size $224\times224$. Predictions from these global and local image views are aggregated to an overall score by a fully-connected layer. Similar to Murray et al.~\cite{murray2012ava}, in~\cite{lu2015rating} images are also categorized to low and high aesthetics based on mean human ratings. A regression loss and an AlexNet inspired  architecture is used in~\cite{kao2015visual} to predict the mean scores. In a similar approach to~\cite{kao2015visual}, Bin et al.~\cite{jin2016image} fine-tune a VGG network~\cite{simonyan2014very} to learn the human ratings of the AVA dataset. They use a regression framework to predict the histogram of ratings. A recent method by Zheng et al.~\cite{zeng2017probabilistic} retrains AlexNet and ResNet CNNs to predict quality of photos. More recently,~\cite{mai2016composition} uses an adaptive spatial pooling to allow for feeding multiple scales of the input image with fixed size aspect ratios to their CNN. This work presents a multi-net (each network a pre-trained VGG) approach which extracts features at multiple scales, and uses a scene aware aggregation layer to combine predictions of the sub-networks. Similarly, Ma et al.~\cite{ma2017lamp} propose a layout-aware framework in which a saliency map is used to select patches with highest impact on predicted aesthetic score. Overall, none of these methods reported correlation of their predictions with respect to ground truth ratings. Recently, Kong et al. in~\cite{kong2016photo} proposed a method to aesthetically rank photos by training on AVA with a rank-based loss function. They trained an AlexNet-based CNN to learn the difference of the aesthetic scores from two input images, and as a result, indirectly optimize for rank correlation. To the best of our knowledge, \cite{kong2016photo} is the only work that performed a correlation evaluation against AVA ratings.

\begin{figure*}[!t]
\vspace{-0 mm}
\begin{center}
\subfigure{
\includegraphics*[viewport=5 1 535 400, scale=0.275]{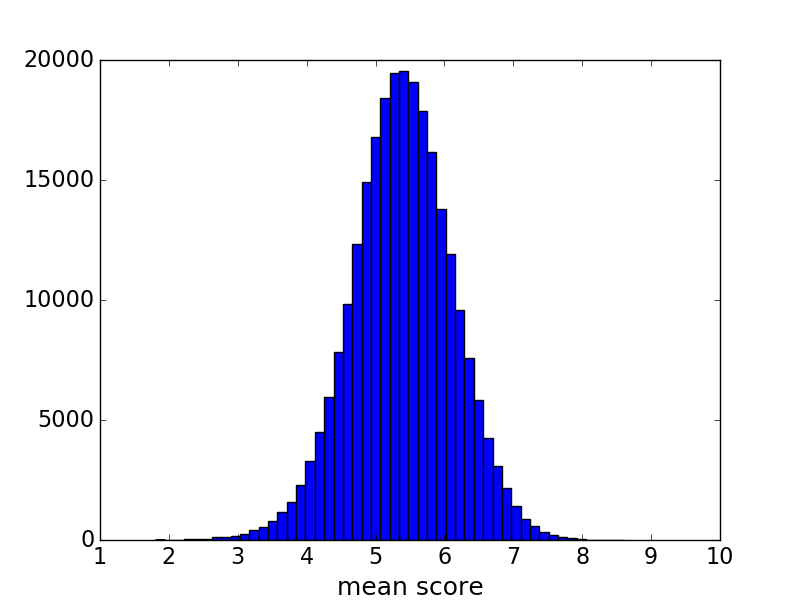} }
\subfigure{
\includegraphics*[viewport=5 1 535 400, scale=0.275]{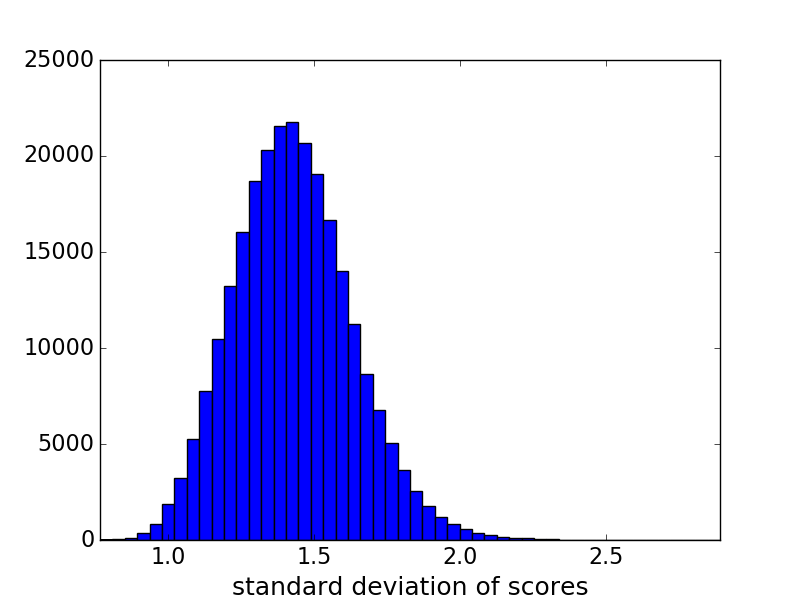} }
\subfigure{
\includegraphics*[viewport=5 1 535 400, scale=0.275]{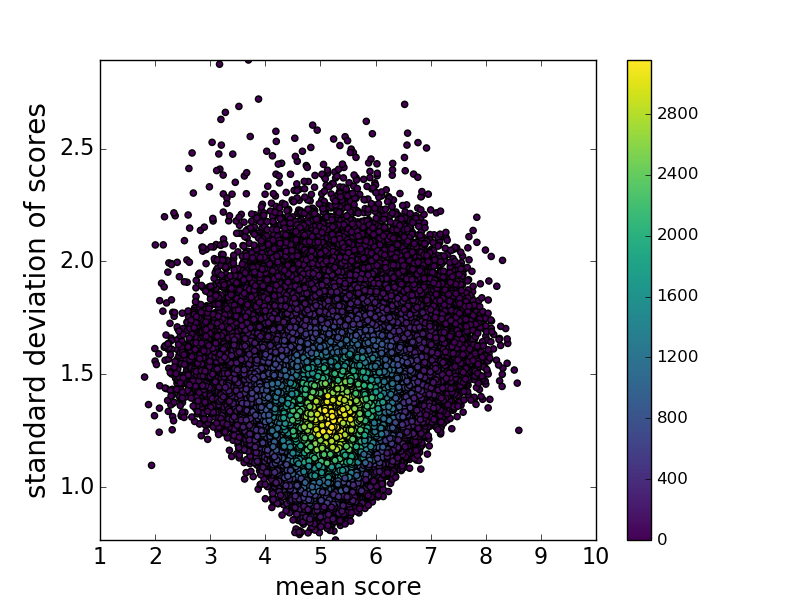} }
\end{center}
\vspace{-5 mm}
{\caption{Histograms of ratings from AVA dataset\cite{murray2012ava}. Left: Histogram of mean scores. Middle: Histogram of standard deviations. Right: Joint histogram of the mean and standard deviation. \label{fig:ava_histograms}}}
\vspace{-0 mm}
\end{figure*}

 \begin{figure*}[!t]
\vspace{-5 mm}
\begin{center}
\subfigure[\scriptsize $6.36\:(\pm1.04)$]{
\includegraphics*[viewport=5 1 475 700, scale=0.12]{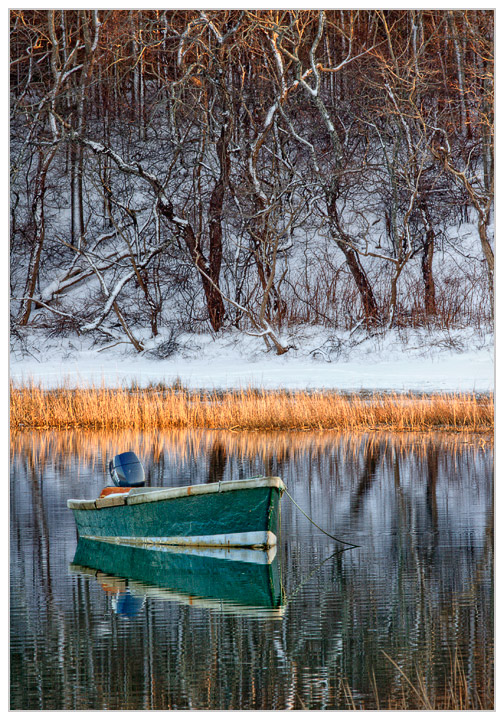} }
\subfigure[\scriptsize $7.84\:(\pm2.08)$]{
\includegraphics*[viewport=5 1 250 200, scale=0.53]{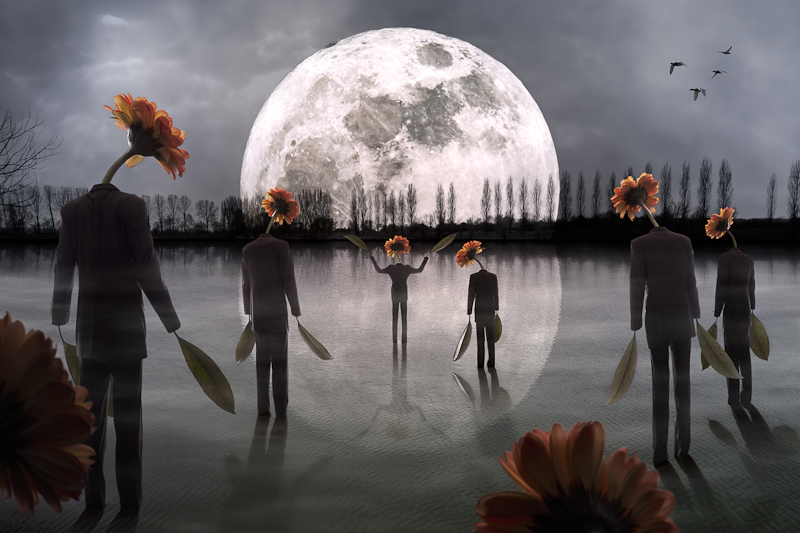} }
\subfigure[\scriptsize $2.62\:(\pm2.15)$]{
\includegraphics*[viewport=5 1 260 200, scale=0.44]{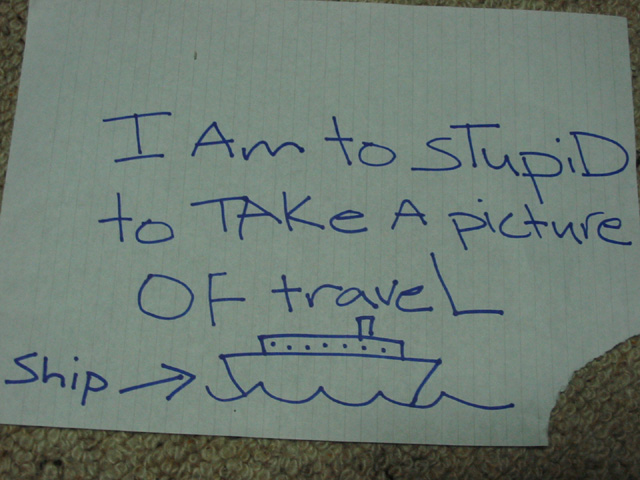} }
\subfigure[\scriptsize $3.12\:(\pm1.28)$]{
\includegraphics*[viewport=5 1 580 420, scale=0.2]{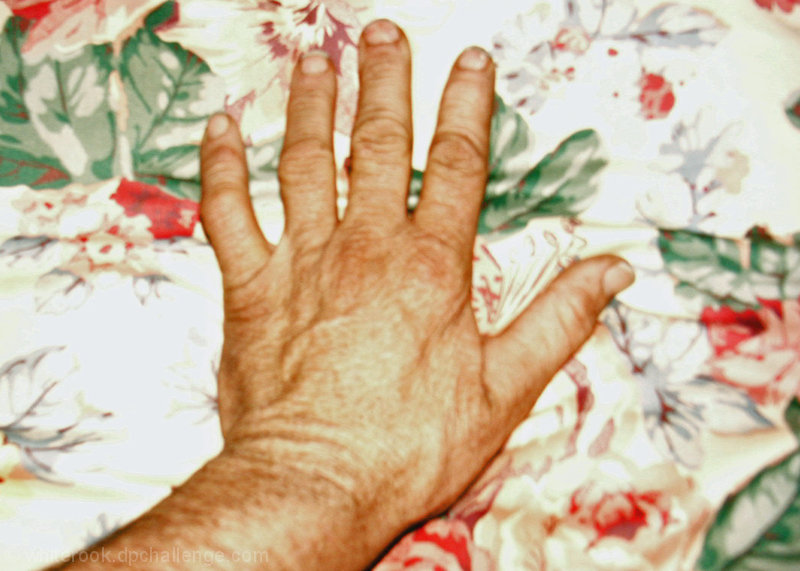} }
\end{center}
\vspace{-5 mm}
{\caption{Some example images from AVA dataset\cite{murray2012ava} with quality score $\mu(\pm\sigma)$, where $\mu$ and $\sigma$ represent mean and standard deviation of score, respectively. (a) high aesthetics and low unconventionality (challenge name: ``\textit{Best of 2007}", $\mu=6.36$, $\sigma =1.04$), (b) high aesthetics and high unconventionality (challenge name: ``\textit{Extreme super moon}", $\mu =7.84$,  $\sigma=2.08$), (c) low aesthetics and high unconventionality (challenge name: ``\textit{Travel}", $\mu=2.62$, $\sigma=2.15$), (d) low aesthetics and low unconventionality (challenge name: ``\textit{Pieces of the human form}", $\mu=3.12$, $\sigma =1.28$). \label{fig:ava_photos}}}
\vspace{-0 mm}
\end{figure*}

 \subsection{Our Contributions}
In this work, we introduce a novel approach to predict both technical and aesthetic qualities of images. We show that models with the same CNN architecture, trained on different datasets, lead to state-of-the-art performance for both tasks. Since we aim for predictions with higher correlation with human ratings, instead of classifying images to low/high score or regressing to the mean score, the distribution of ratings are predicted as a histogram. To this end, we use the squared EMD (earth mover's distance) loss proposed in~\cite{hou2016squared}, which shows a performance boost in classification with ordered classes. Our experiments show that this approach also leads to more accurate prediction of the mean score. Also, as shown in aesthetic assessment case~\cite{murray2012ava}, non-conventionality of images is directly related to score standard deviations. Our proposed paradigm allows for predicting this metric as well.

It has recently been shown that perceptual quality predictors can be used as learning loss to train image enhancement models \cite{zhang2018unreasonable, talebi2017learned}. Similarly, image quality predictors can be used to adjust parameters of enhancement techniques \cite{gu2016analysis}. In this work we use our quality assessment technique to effectively tune parameters of image denoising and tone enhancement operators to produce perceptually superior results. 

This paper begins with reviewing three widely used datasets for quality assessment. Then, our proposed method is explained in more detail. Finally, performance of this work is quantified and compared to the existing methods. 

 \subsection{A Large-Scale Database for Aesthetic Visual Analysis (AVA) \cite{murray2012ava}}
 The AVA dataset contains about 255,000 images, rated based on aesthetic qualities by amateur photographers\footnote{AVA images are obtained from \hyperref[www.dpchallenge.com]{www.dpchallenge.com}, which is an on-line community for amateur photographers.}. Each photo is scored by an average of 200 people in response to photography contests. Each image is associated to a single challenge theme, with nearly 900 different contests in the AVA. The image ratings range from 1 to 10, with 10 being the highest aesthetic score associated to an image. Histograms of AVA ratings are shown in Fig.~\ref{fig:ava_histograms}. As can be seen, mean ratings are concentrated around the overall mean score ($\approx$5.5). Also, ratings of roughly half of the photos in AVA dataset have a standard deviation greater than 1.4. As pointed out in~\cite{murray2012ava}, presumably images with high score variance tend to be subject to interpretation, whereas images with low score variance seem to represent conventional styles or subject matter. A few examples with ratings associated with different levels of aesthetic quality and unconventionality are illustrated in Fig.~\ref{fig:ava_photos}. It seems that aesthetic quality of a photograph can be represented by the mean score, and unconventionality of it closely correlates to the score deviation. Given the distribution of AVA scores, typically, training a model on AVA data results in predictions with small deviations around the overall mean (5.5). 
 
It is worth mentioning that the joint histogram in Fig.~\ref{fig:ava_histograms} shows higher deviations for very low/high ratings (compared to the overall mean 5.5, and mean standard deviation 1.43). In other words, divergence of opinion is more consistent in AVA images with extreme aesthetic qualities. As discussed in~\cite{murray2012ava}, distribution of ratings with mean value between 2 and 8 can be closely approximated by Gaussian functions, and highly skewed ratings can be modeled by Gamma distributions.

 \begin{figure*}[!t]
\vspace{-0 mm}
\begin{center}
\subfigure{
\includegraphics*[viewport=5 1 535 400, scale=0.275]{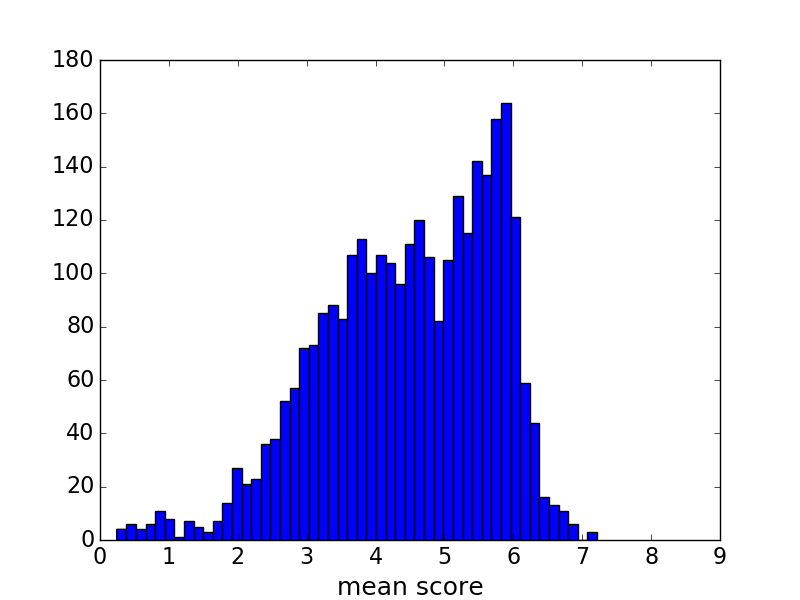} }
\subfigure{
\includegraphics*[viewport=5 1 535 400, scale=0.275]{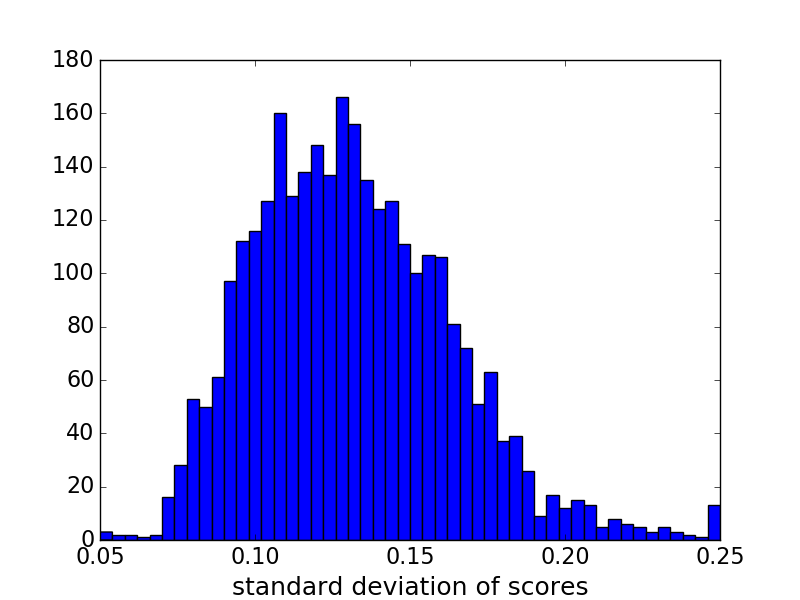} }
\subfigure{
\includegraphics*[viewport=5 1 535 400, scale=0.275]{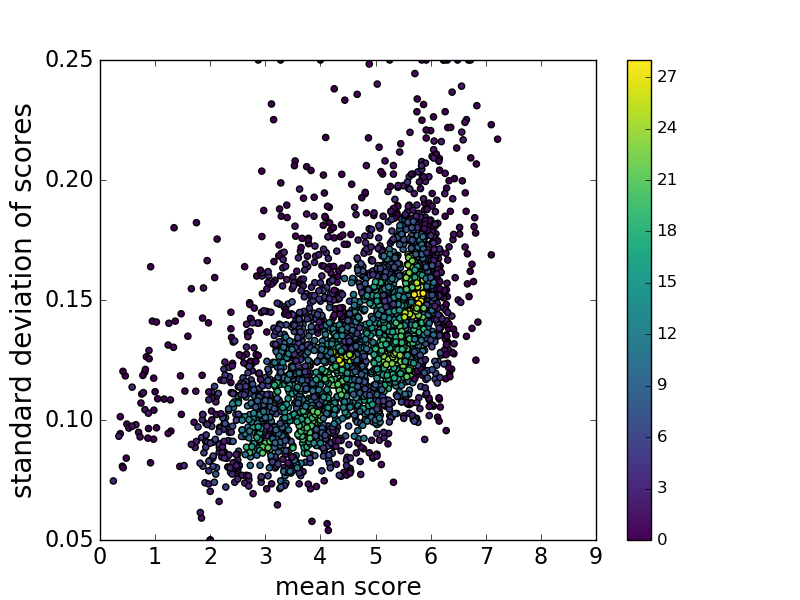} }
\end{center}
\vspace{-5 mm}
{\caption{Histograms of ratings from TID2013 dataset\cite{ponomarenko2013color}. Left: Histogram of mean scores. Middle: Histogram of standard deviations. Right: Joint histogram of the mean and standard deviation. \label{fig:tid_histograms}}}
\vspace{-0 mm}
\end{figure*}

 \subsection{Tampere Image Database 2013 (TID2013) \cite{ponomarenko2013color}}
TID2013 is curated for evaluation of full-reference perceptual image quality. It contains 3000 images, from 25 reference (clean) images (Kodak images~\cite{kodak_dataset}), 24 types of distortions with 5 levels for each distortion. This leads to 120 distorted images for each reference image; including different types of distortions such as compression artifacts, noise, blur and color artifacts. 

Human ratings of TID2013 images are collected through a forced choice experiment, where observers select a better image between two distorted choices. Set up of the experiment allows raters to view the reference image while making a decision. In each experiment, every distorted image is used in 9 random pairwise comparisons. The selected image gets one point, and other image gets zero points. At the end of the experiment, sum of the points is used as the quality score associated with an image (this leads to scores ranging from 0 to 9). To obtain the overall mean scores, total of 985 experiments are carried out.

\begin{figure*}[!t]
\vspace{-0 mm}
\begin{center}
\subfigure[\scriptsize clean image]{
\includegraphics*[viewport=1 1 512 384, scale=0.28]{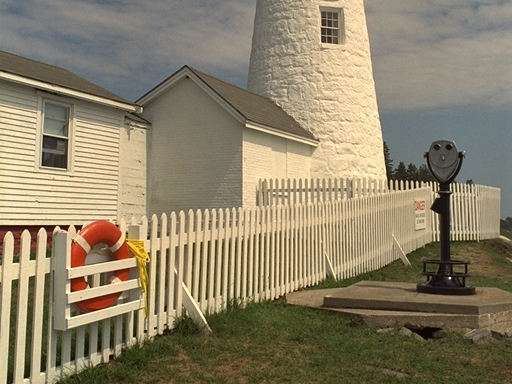} }
\subfigure[\scriptsize $5.73\:(\pm0.15)$]{
\includegraphics*[viewport=1 1 512 384, scale=0.28]{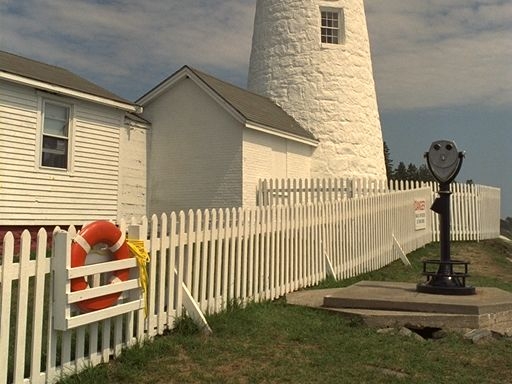} }
\subfigure[\scriptsize $5.47\:(\pm0.11)$]{
\includegraphics*[viewport=1 1 512 384, scale=0.28]{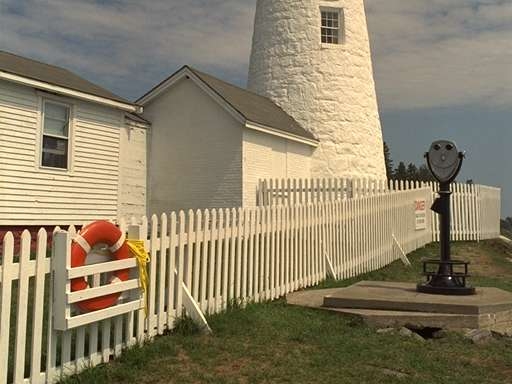} }
\subfigure[\scriptsize $4.86\:(\pm0.11)$ ]{
\includegraphics*[viewport=1 1 512 384, scale=0.28]{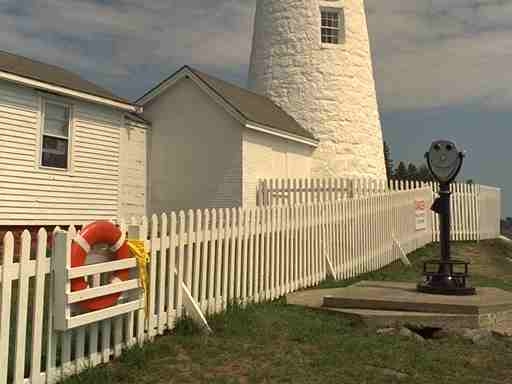} }
\subfigure[\scriptsize $3.0\:(\pm0.11)$ ]{
\includegraphics*[viewport=1 1 512 384, scale=0.28]{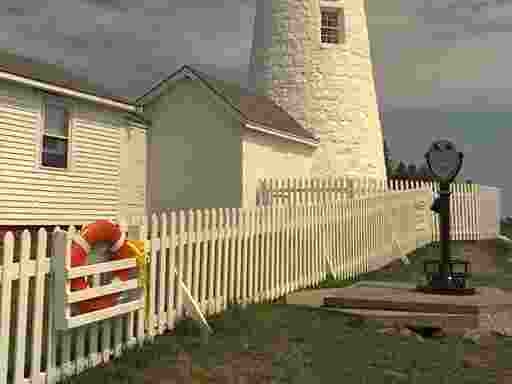} }
\subfigure[\scriptsize $1.66\:(\pm0.16)$]{
\includegraphics*[viewport=1 1 512 384, scale=0.28]{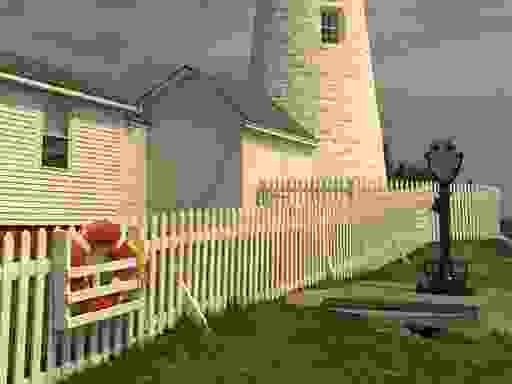} }
\end{center}
\vspace{-0 mm}
{\caption{JPEG artifact example images from TID2013 dataset\cite{ponomarenko2013color} with quality score $\mu(\pm\sigma)$, where $\mu$ and $\sigma$ represent mean and standard deviation of score, respectively. Clean image and 5 levels of JPEG compression artifacts are shown here. (a) clean image, (b) compression artifact level 1, $\mu=5.73$, $\sigma=0.15$, (c) compression artifact level 2, $\mu=5.47$, $\sigma=0.11$, (d) compression artifact level 3, $\mu=4.86$, $\sigma=0.11$, (e) compression artifact level 4,  $\mu=3.0$, $\sigma=0.11$, (f) compression artifact level 5, $\mu=1.66$, $\sigma=0.16$. \label{fig:tid_photos1}}}
\vspace{-0 mm}
\end{figure*}

\begin{figure*}[!t]
\vspace{-0 mm}
\begin{center}
\subfigure[\scriptsize clean image]{
\includegraphics*[viewport=1 1 512 384, scale=0.28]{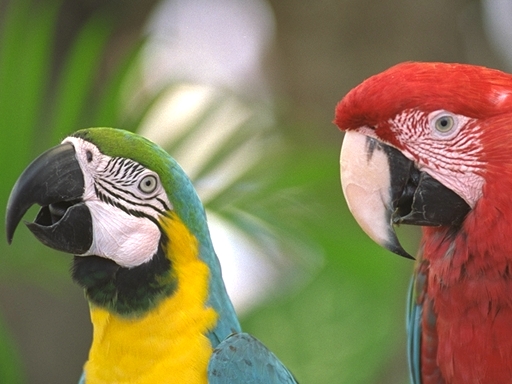} }
\subfigure[\scriptsize $5.67\:(\pm0.10)$]{
\includegraphics*[viewport=1 1 512 384, scale=0.28]{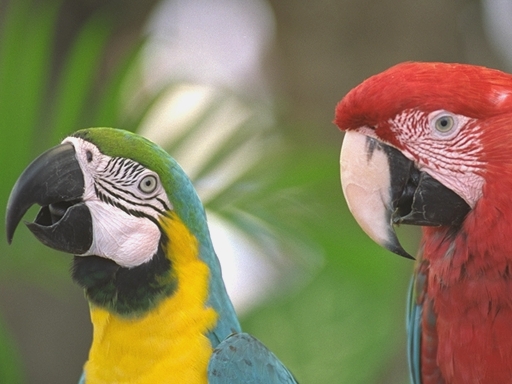} }
\subfigure[\scriptsize $6.80\:(\pm0.18)$]{
\includegraphics*[viewport=1 1 512 384, scale=0.28]{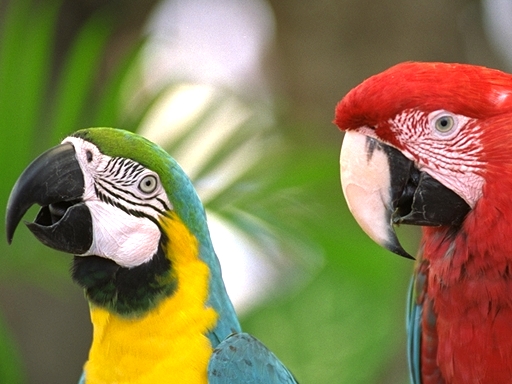} }
\subfigure[\scriptsize $4.83\:(\pm0.16)$]{
\includegraphics*[viewport=1 1 512 384, scale=0.28]{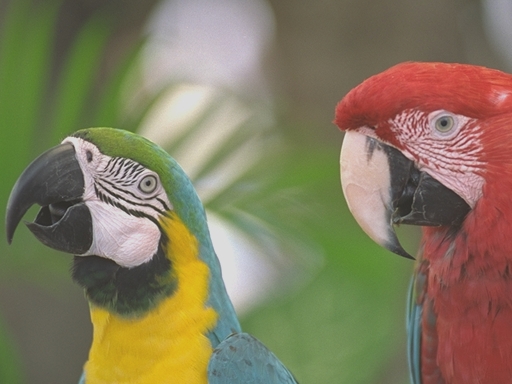} }
\subfigure[\scriptsize $6.69\:(\pm0.29)$]{
\includegraphics*[viewport=1 1 512 384, scale=0.28]{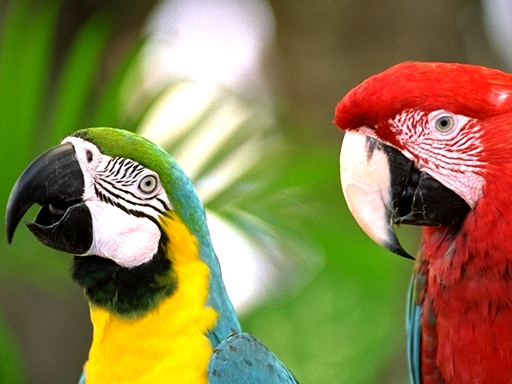} }
\subfigure[\scriptsize $3.88\:(\pm0.18)$]{
\includegraphics*[viewport=1 1 512 384, scale=0.28]{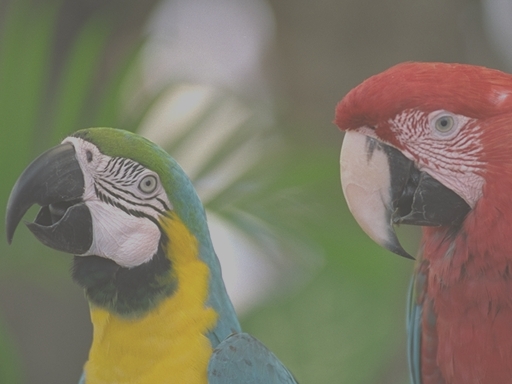} }
\end{center}
\vspace{-0 mm}
{\caption{Some example images from TID2013 dataset\cite{ponomarenko2013color} with quality score $\mu(\pm\sigma)$, where $\mu$ and $\sigma$ represent mean and standard deviation of score, respectively. Clean image and 5 levels of contrast change distortions are shown here. (a) clean image, (b) contrast change distortion of level 1, $\mu=5.67$, $\sigma =0.10$, (c) contrast change distortion of level 2, $\mu=6.80$, $\sigma=0.18$, (d) contrast change distortion of level 3, $\mu=4.83$, $\sigma =0.16$, (e) contrast change distortion of level 4, $\mu=6.69$, $\sigma=0.29$, (f) contrast change distortion of level 5, $\mu=3.88$, $\sigma=0.18$. \label{fig:tid_photos2}}}
\vspace{-0 mm}
\end{figure*}

Mean and standard deviation of TID2013 ratings are shown in Fig.~\ref{fig:tid_histograms}. As can be seen in Fig.~\ref{fig:tid_histograms}(c), the mean and score deviation values are weakly correlated. A few images from TID2013 are illustrated in Fig.~\ref{fig:tid_photos1} and Fig.~\ref{fig:tid_photos2}. All five levels of JPEG compression artifacts and the respective ratings are illustrated in Fig.~\ref{fig:tid_photos1}. Evidently higher distortion level leads to lower mean score\footnote{This is a quite consistent trend for most of the other distortions too (namely noise, blur and color distortions). However, in case of the contrast change (Fig.~\ref{fig:tid_photos2}), this trend is not obvious. This is due to the order of contrast compression/stretching from level 1 to level 5)}. Effect of contrast compression/stretching distortion on the human ratings is demonstrated in Fig.~\ref{fig:tid_photos2}. Interestingly, stretch of contrast (Fig.~\ref{fig:tid_photos2}(c) and Fig.~\ref{fig:tid_photos2}(e)) leads to relatively higher perceptual quality.

 \subsection{LIVE In the Wild Image Quality Challenge Database \cite{ghadiyaram2016massive}}
\begin{figure*}[!t]
\vspace{-0 mm}
\begin{center}
\subfigure{
\includegraphics*[viewport=5 1 535 400, scale=0.275]{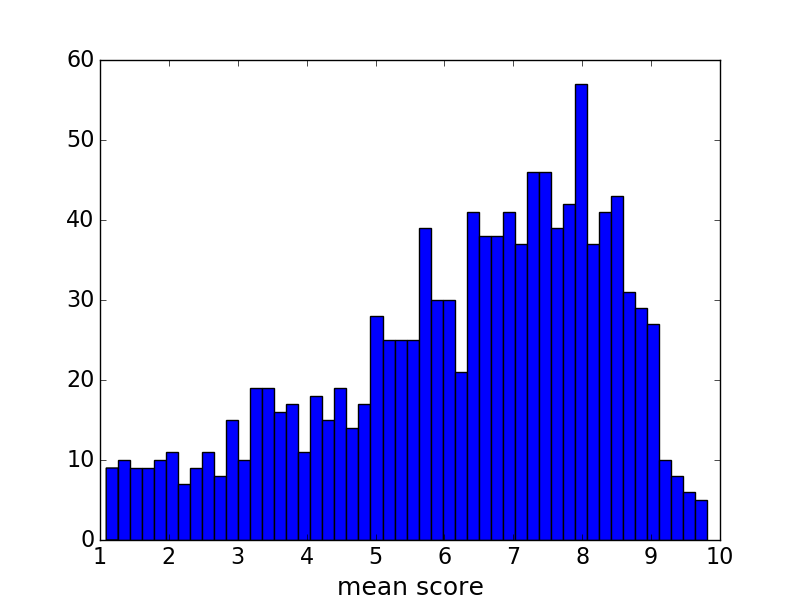} }
\subfigure{
\includegraphics*[viewport=5 1 535 400, scale=0.275]{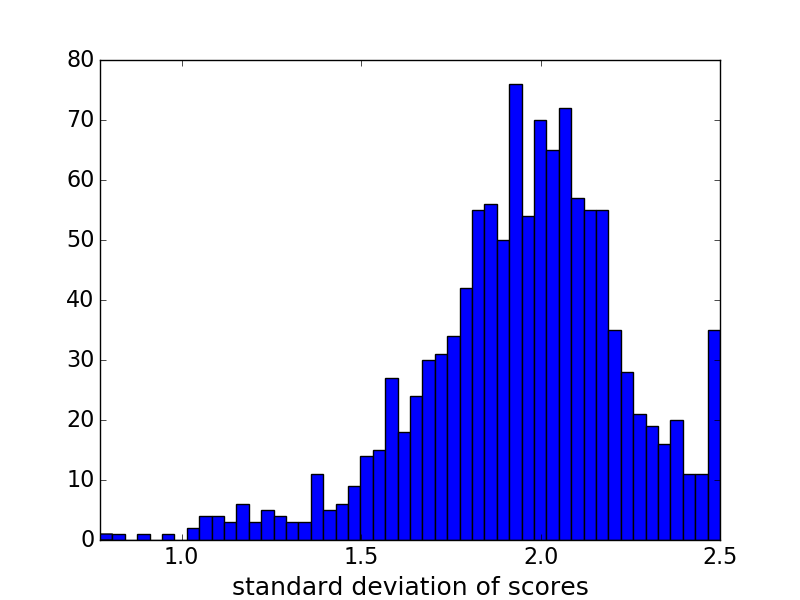} }
\subfigure{
\includegraphics*[viewport=5 1 535 400, scale=0.275]{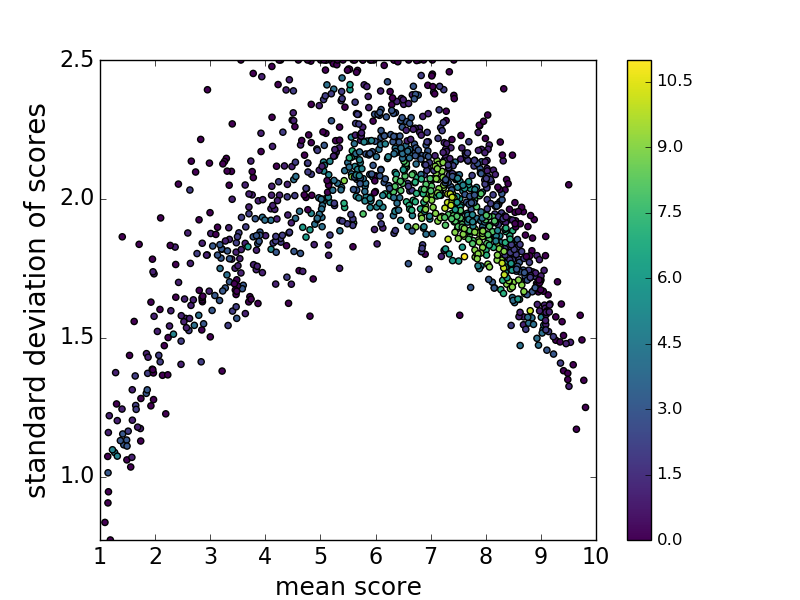} }
\end{center}
\vspace{-5 mm}
{\caption{Histograms of ratings from LIVE dataset~\cite{ghadiyaram2016massive}. Left: Histogram of mean scores. Middle: Histogram of standard deviations. Right: Joint histogram of the mean and standard deviation. Note that LIVE scores are scaled to [1,10]. \label{fig:live_histograms}}}
\vspace{-0 mm}
\end{figure*}

\begin{figure*}[!t]
\vspace{-0 mm}
\begin{center}
\subfigure[\scriptsize $9.99\:(\pm1.22)$]{
\includegraphics*[scale=0.24]{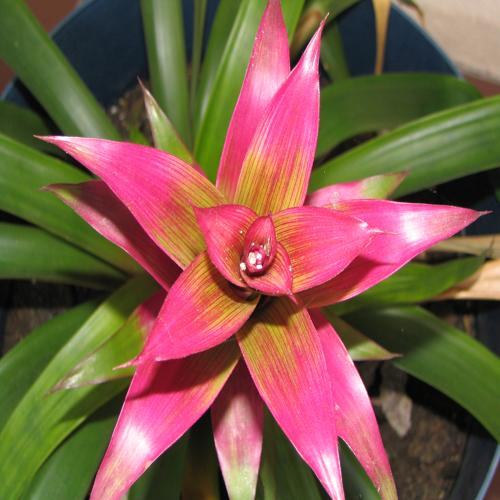} }
\subfigure[\scriptsize $9.35\:(\pm1.49)$]{
\includegraphics*[scale=0.24]{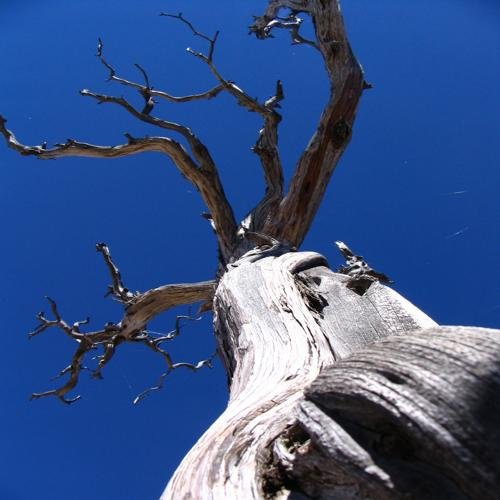} }
\subfigure[\scriptsize $8.29\:(\pm1.99)$]{
\includegraphics*[scale=0.24]{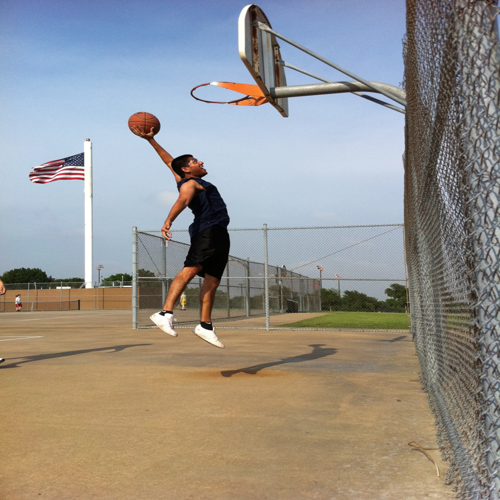} }
\subfigure[\scriptsize $3.50\:(\pm1.69)$]{
\includegraphics*[scale=0.24]{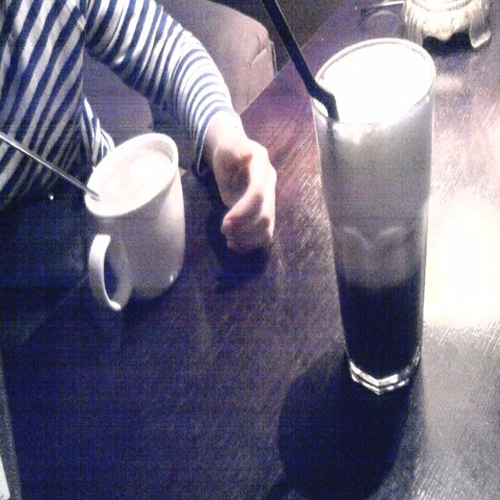} }
\subfigure[\scriptsize $2.33\:(\pm1.51)$]{
\includegraphics*[scale=0.24]{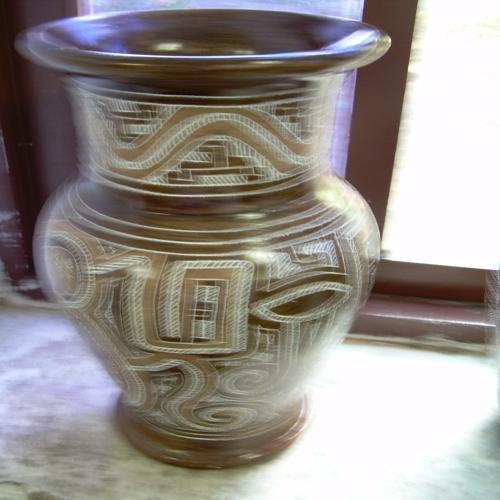} }
\subfigure[\scriptsize $1.95\:(\pm1.39)$]{
\includegraphics*[scale=0.24]{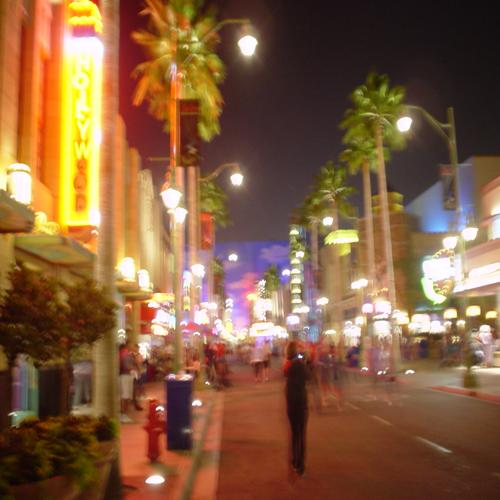} }
\end{center}
\vspace{-0 mm}
{\caption{Some example images from LIVE dataset~\cite{ghadiyaram2016massive} with quality score $\mu(\pm\sigma)$, where $\mu$ and $\sigma$ represent mean and standard deviation of score, respectively. Note that LIVE scores are scaled to [1,10]. \label{fig:live_photos}}}
\vspace{-0 mm}
\end{figure*}

LIVE dataset contains 1162 photos captured by mobile devices. Each image is rated by an average of 175 unique subjects. Mean and standard deviation of LIVE ratings are shown in Fig.~\ref{fig:live_histograms}. As can be seen in the joint histogram, images that are rated near overall mean score show higher standard deviation.  A few images from LIVE dataset are illustrated in Fig.~\ref{fig:live_photos}. It is worth noting that in this paper, LIVE scores are scaled to $[1,\:10]$.

Unlike AVA, which includes distribution of ratings for each image, TID2013 and LIVE only provide mean and standard deviation of the opinion scores. Since our proposed method requires training on score probabilities, the score distributions are approximated through maximum entropy optimization~\cite{cover2012elements}.

The rest of the paper is organized as follows. In Section~\ref{sec:proposed}, a detailed explanation of the proposed method is described. Next, in Section\ref{sec:results}, applications of our algorithm in ranking photos and image enhancement are exemplified. We also provide details of our implementation. Finally, this paper is concluded in Section\ref{sec:conclusion}.

\section{Proposed Method}
\label{sec:proposed}

\begin{figure*}[!t]
\vspace{-0 mm}
\begin{center}
\subfigure{
\includegraphics*[viewport=1 1 820 260, scale=0.3]{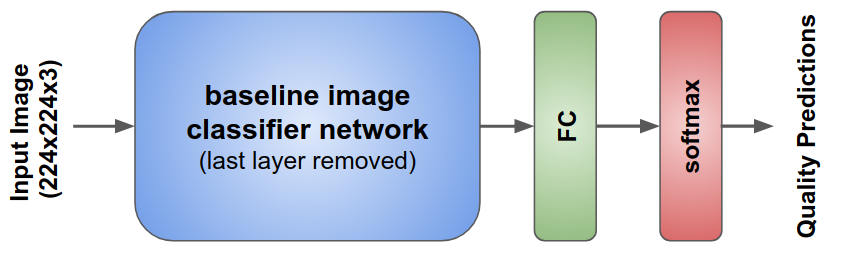}}
\end{center}
\vspace{-4 mm}
{\caption{Modified baseline image classifier network used in our framework. Last layer of classifier network is replaced by a fully-connected layer to output 10 classes of quality scores. Baseline network weights are initialized by training on ImageNet dataset~\cite{krizhevsky2012imagenet}, and the added fully-connected weights are initialized randomly. \label{fig:cnn_architecture}}}
\vspace{-0 mm}
\end{figure*}

Our proposed quality and aesthetic predictor stands on image classifier architectures. More explicitly, we explore a few different classifier architectures such as VGG16~\cite{simonyan2014very}, Inception-v2~\cite{szegedy2016rethinking}, and MobileNet~\cite{howard2017mobilenets} for image quality assessment task. VGG16 consists of 13 convolutional and 3 fully-connected layers. Small convolution filters of size $3\times3$ are used in the deep VGG16 architecture~\cite{simonyan2014very}. Inception-v2~\cite{szegedy2016rethinking} is based on the Inception module~\cite{ioffe2015batch} which allows for parallel use of convolution and pooling operations. Also, in the Inception architecture, traditional fully-connected layers are replaced by average pooling, which leads to a significant reduction in number of parameters. MobileNet~\cite{howard2017mobilenets} is an efficient deep CNN, mainly designed for mobile vision applications. In this architecture, dense convolutional filters are replaced by separable filters. This simplification results in smaller and faster CNN models. 

We replaced the last layer of the baseline CNN with a fully-connected layer with 10 neurons followed by soft-max activations (shown in Fig.~\ref{fig:cnn_architecture}). Baseline CNN weights are initialized by training on the ImageNet dataset~\cite{krizhevsky2012imagenet}, and then an end-to-end training on quality assessment is performed. In this paper, we discuss performance of the proposed model with various baseline CNNs.

In training, input images are rescaled to $256\times256$, and then a crop of size $224\times224$ is randomly extracted. This lessens potential over-fitting issues, especially when training on relatively small datasets (e.g.\ TID2013). It is worth noting that we also tried training with random crops without rescaling. However, results were not compelling. This is due to the inevitable change in image composition. Another random data augmentation in our training process is horizontal flipping of the image crops.

Our goal is to predict the distribution of ratings for a given image. Ground truth distribution of human ratings of a given image can be expressed as an empirical probability mass function $\textbf{p} = [p_{s_1},\ldots,p_{s_N}]$ with $s_1 \leq s_i \leq s_N$, where $s_i$ denotes the $i$th score bucket, and $N$ denotes the total number of score buckets. In both AVA and TID2013 datasets $N=10$,  in AVA, $s_1=1$ and $s_N=10$, and in TID $s_1=0$ and $s_N=9$. Since $\sum_{i=1}^{N} p_{s_i} = 1$, $p_{s_i}$ represents the probability of a quality score falling in the $i$th bucket. Given the distribution of ratings as $\textbf{p}$, mean quality score is defined as $\mu = \sum_{i=1}^{N} s_{i} \times p_{s_i}$, and standard deviation of the score is computed as $\sigma = (\sum_{i=1}^{N} (s_{i}-\mu)^2 \times p_{s_{i}})^{1/2}$. As discussed in the previous section, one can qualitatively compare images by mean and standard deviation of scores.

Each example in the dataset consists of an image and its ground truth (user) ratings $\textbf{p}$. Our objective is to find the probability mass function $\widehat{\textbf{p}}$ that is an accurate estimate of $\textbf{p}$. Next, our training loss function is discussed.

\subsection{Loss Function}
Soft-max cross-entropy is widely used as training loss in classification tasks. This loss can be represented as $\sum_{i=1}^{N} -p_{s_i} \log(\widehat{p}_{s_i})$ (where $\widehat{p}_{s_i}$ denotes estimated probability of $i$th score bucket) to maximize predicted probability of the correct labels. However, in the case of ordered-classes (e.g.\ aesthetic and quality estimation), cross-entropy loss lacks the inter-class relationships between score buckets. One might argue that ordered-classes can be represented by a real number, and consequently, can be learned through a regression framework. Yet, it has been shown that for ordered classes, the classification frameworks can outperform regression models~\cite{hou2016squared, golik2013cross}. Hou et al.~\cite{hou2016squared} show that training on datasets with intrinsic ordering between classes can benefit from EMD-based losses. These loss functions penalize mis-classifications according to class distances.

For image quality ratings, classes are inherently ordered as $s_1 < \dots  < s_N$, and $r-$norm distance between classes is defined as $\|s_i - s_j\|_r$, where $1 \leq i, j \leq N$. EMD is defined as the minimum cost to move the mass of one distribution to another. Given the ground truth and estimated probability mass functions $\textbf{p}$ and $\widehat{\textbf{p}}$, with $N$ ordered classes of distance $\|s_i - s_j\|_r$, the normalized Earth Mover's Distance can be expressed as~\cite{levina2001earth}:

\begin{equation}
\label{eqn:emd}
\mbox{EMD}(\textbf{p}, \widehat{\textbf{p}}) = \left( \frac{1}{N} \sum_{k=1}^{N} |\mbox{CDF}_{\textbf{p}}(k) - \mbox{CDF}_{\widehat{\textbf{p}}}(k)|^r \right)^{1/r}
\end{equation}
where $\mbox{CDF}_{\textbf{p}}(k)$ is the cumulative distribution function as $\sum_{i=1}^{k} \textbf{p}_{s_i}$. It is worth noting that this closed-form solution requires both distributions to have equal mass as $\sum_{i=1}^{N} \textbf{p}_{s_i} = \sum_{i=1}^{N} \widehat{\textbf{p}}_{s_i}$. As shown in Fig.~\ref{fig:cnn_architecture}, our predicted quality probabilities are fed to a soft-max function to guarantee that $\sum_{i=1}^{N} \widehat{\textbf{p}}_{s_i} = 1$. Similar to~\cite{hou2016squared}, in our training framework, $r$ is set as 2 to penalize the Euclidean distance between the CDFs. $r=2$ allows easier optimization when working with gradient descent.

\section{Experimental Results}
\label{sec:results}
\vspace{0 mm}

We train two separate models for aesthetics and technical quality assessment on AVA, TID2013, and LIVE. For each case, we split each dataset into train and test sets, such that 20\% of the data is used for testing. In this section, performance of the proposed models on the test sets are discussed and compared to the existing methods. Then, applications of the proposed technique in photo ranking and image enhancement are explored. Before moving forward, details of our implementation are explained.

The CNNs presented in this paper are implemented using TensorFlow~\cite{abadi2016tensorflow, abadi2016tensorflow2}. The baseline CNN weights are initialized by training on ImageNet~\cite{krizhevsky2012imagenet}, and the last fully-connected layer is randomly initialized. The weight and bias momentums are set to 0.9, and a dropout rate of 0.75 is applied on the last layer of the baseline network. The learning rate of the baseline CNN layers and the last fully-connected layers are set as $3\times10^{-7}$ and $3\times 10^{-6}$, respectively. We observed that setting a low learning rate on baseline CNN layers results in easier and faster optimization when using stochastic gradient descent. Also, after every 10 epochs of training, an exponential decay with decay factor 0.95 is applied to all learning rates.

\subsection{Performance Comparisons}
\label{sec:performance}
\vspace{0 mm}

Accuracy, correlation and EMD values of our evaluations on the aesthetic assessment model on AVA are presented in Table~\ref{tab:ava_comp}. Most methods in Table~\ref{tab:ava_comp} are designed to perform binary classification on the aesthetic scores, and as a result, only accuracy evaluations of two-class quality categorization are reported. In this binary classification, predicted mean scores are compared to $5$ as cut-off score. Images with predicted scores above the cut-off score are categorized as high quality. In two-class aesthetic categorization task, results from~\cite{ma2017lamp}, and NIMA(Inception-v2) show the highest accuracy. Also, in terms of rank correlation, NIMA(VGG16) and NIMA(Inception-v2) outperform~\cite{kong2016photo}. \emph{NIMA is much cheaper}: \cite{ma2017lamp} applies multiple VGG16 nets on image patches to generate a single quality score, whereas computational complexity of NIMA(Inception-v2) is roughly one pass of Inception-v2 (see Table~\ref{tab:time_comp}).

Our technical quality assessment model on TID2013 is compared to other existing methods in Table~\ref{tab:tid_comp}. While most of these methods regress to the mean opinion score, our proposed technique predicts the distribution of ratings, as well as mean opinion score. Correlation between ground truth and results of NIMA(VGG16) are close to the state-of-the-art results in~\cite{xu2016blind} and~\cite{bianco2016use}. It is worth highlighting that Bianco et al.~\cite{bianco2016use} feed multiple image crops to a deep CNN, whereas our method takes only the rescaled image.

\begin{figure}[!t]
\vspace{0 mm}
\begin{center}
\subfigure{
\includegraphics*[viewport=10 10 560 430, scale=0.21]{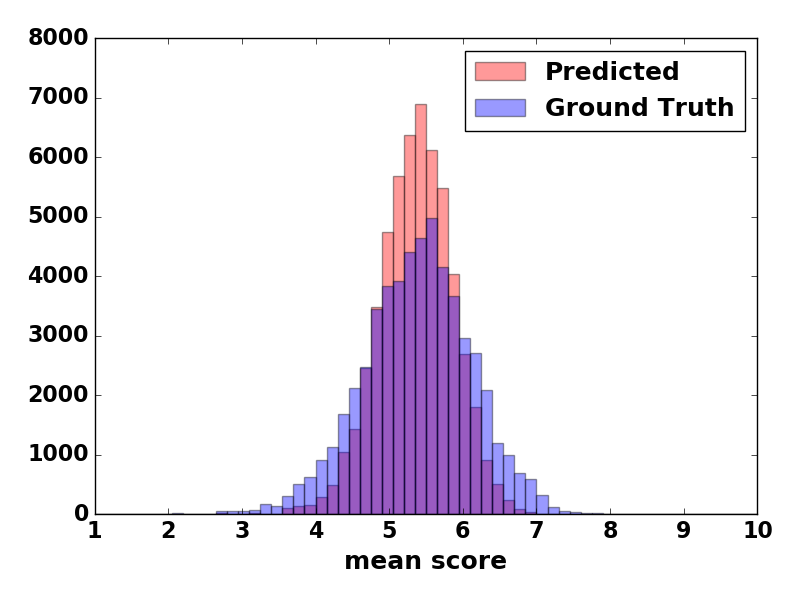} }
\subfigure{
\includegraphics*[viewport=10 10 560 430, scale=0.21]{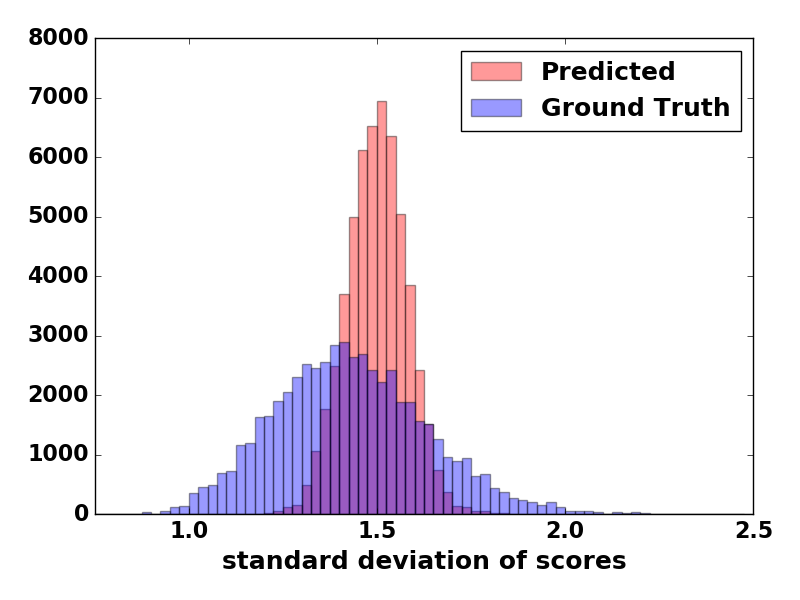} }
\end{center}
\vspace{-3 mm}
{\caption{Histograms of the ground truth and predicted scores using NIMA(Inception-v2) applied on our AVA test set. Left: histograms of mean scores. Right: histograms of standard deviations. \label{fig:ava_eval_histograms}}}
\vspace{-0 mm}
\end{figure}

The predicted distributions of AVA  scores are presented in Fig. \ref{fig:ava_eval_histograms}. We used NIMA(Inception-v2) model to predict the ground truth scores from our AVA test set. As can be seen, distribution of the ground truth mean scores is closely predicted by NIMA. However, predicting distribution of the ground truth standard deviations is a more challenging task. As we discussed previously, unconventionality of subject matter or style has a direct impact on score standard deviations.

\begin{table*}[!t]
\vspace{12 mm}
\begin{center}
\captionsetup{width=0.57\textwidth}
\caption{Performance of the proposed method with various architectures in predicting AVA quality ratings~\cite{murray2012ava} compared to the state-of-the-art. Reported accuracy values are based on classification of photos to two classes (column 2). LCC (linear correlation coefficient) and SRCC (Spearman's rank correlation coefficient) are computed between predicted and ground truth mean scores (column 3 and 4) and standard deviation of scores (column 5 and 6). EMD measures closeness of the predicted and ground truth rating distributions with $r=1$ in Eq. \ref{eqn:emd}. The accuracy, LCC, and SROC values are in $\pm0.3$,  $\pm0.005$, and $\pm0.004$ within $95\%$ confidence, respectively.}
\begin{tabular}{@{} *7l @{}}    \toprule
\emph{Model} & \emph{Accuracy} & \emph{LCC} & \emph{SRCC} & \emph{LCC} & \emph{SRCC} & \emph{EMD} \\
  & \emph{(2 classes)} & \emph{(mean)} & \emph{(mean)} & \emph{(std.dev)} & \emph{(std.dev)}\\\midrule
Murray et al.~\cite{murray2012ava}    & 66.70\% &  --  &  -- &  --  &  -- &  --   \\ 
Kao et al.~\cite{kao2015visual}    & 71.42\% &  --  &  --  &  --  &  --  &  --  \\ 
Lu et al. \cite{lu2014rapid}    & 74.46\%  & -- &  --  &  --  &  --  &  --  \\ 
Lu et al. \cite{lu2015rating}    & 75.42\%  & -- &  --  &  --  &  --  &  --  \\ 
Kao et al. \cite{kao2016visual}    & 76.58\% & -- &  --  &  --  &  -- &  --   \\ 
Wang et al. \cite{wang2016brain}    & 76.80\% & -- &  --  &  --  &  --  &  --  \\ 
Mai et al. \cite{mai2016composition}    & 77.10\%  & -- &  -- &  --  &  --  &  --  \\ 
Kong et al. \cite{kong2016photo}  &  77.33\% & --  & 0.558 &  --  &  --  &  --  \\
Ma et al. \cite{ma2017lamp} & 81.70\% & -- &  -- &  --  &  --  &  --  \\ \hdashline
NIMA(MobileNet)  & 80.36\%  & 0.518 & 0.510 &  0.152 &  0.137 &  0.081 \\
NIMA(VGG16)  & 80.60\%  & 0.610 & 0.592 &  0.205 &  0.202 &  0.052 \\
NIMA(Inception-v2)  & 81.51\%  & 0.636 & 0.612 &  0.233 &  0.218 &  0.050 \\\bottomrule
 \hline
\end{tabular}
\label{tab:ava_comp}
\end{center}
\end{table*}

\subsection{Cross Dataset Evaluation}
\label{sec:cross_dataset_eval}
\vspace{0 mm}

As a cross validation test, performance of our trained models are measured on other datasets. These results are presented in Table \ref{tab:cross_validation_lcc} and Table \ref{tab:cross_validation_srcc}. We test NIMA(Inception-v2) model trained on AVA, TID2013~\cite{ponomarenko2013color}  and LIVE~\cite{ghadiyaram2016massive} across all three test sets. As can be seen, on average, training on AVA dataset shows the best performance. For instance, training on AVA and testing on LIVE results in $0.552$ and $0.543$ linear and rank correlations, respectively. However, training on LIVE and testing on AVA leads to $0.238$ and $0.2$ linear and rank correlation coefficients. We believe this observation shows that NIMA models trained on AVA can generalize to other test examples more effectively, whereas training on TID2013 results in poor performance on LIVE and AVA test sets. It is worth mentioning that AVA dataset contains roughly 250 times more examples (in comparison to the LIVE dataset), which allows training NIMA models without any significant overfitting.

\subsection{Photo Ranking}
\label{sec:ranking}
\vspace{0 mm}

\begin{figure*}[!t]
\vspace{0 mm}
\begin{center}
\subfigure[\scriptsize 6.38 (7.16)]{
\includegraphics*[scale=0.125]{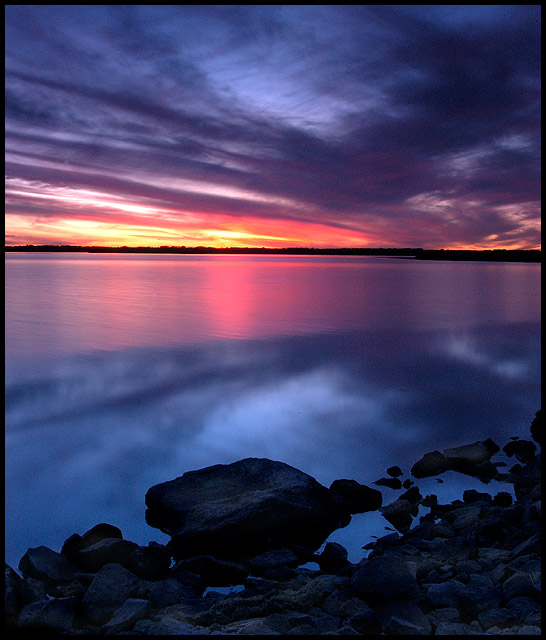} }
\subfigure[\scriptsize 6.24 (6.79)]{
\includegraphics*[scale=0.125]{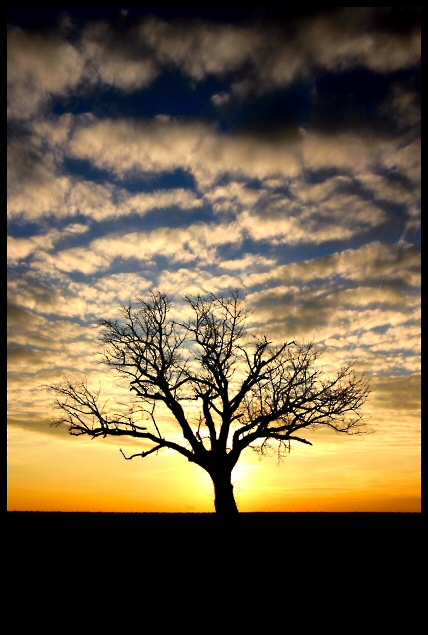} }
\subfigure[\scriptsize 6.22 (6.64)]{
\includegraphics*[scale=0.125]{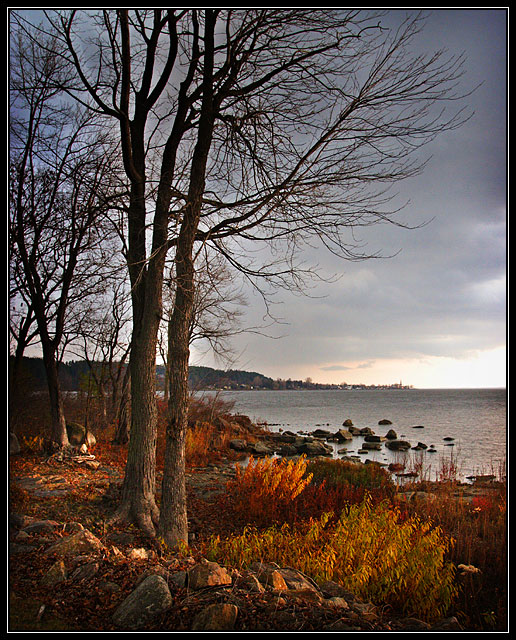} }
\subfigure[\scriptsize 6.16 (6.93)]{
\includegraphics*[scale=0.245]{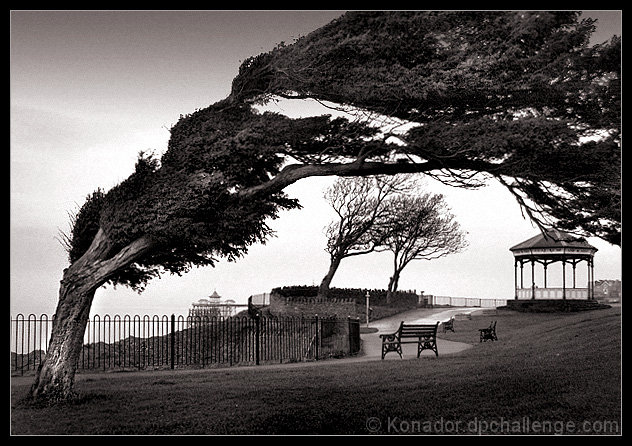} }
\subfigure[\scriptsize 5.92 (6.23)]{
\includegraphics*[scale=0.185]{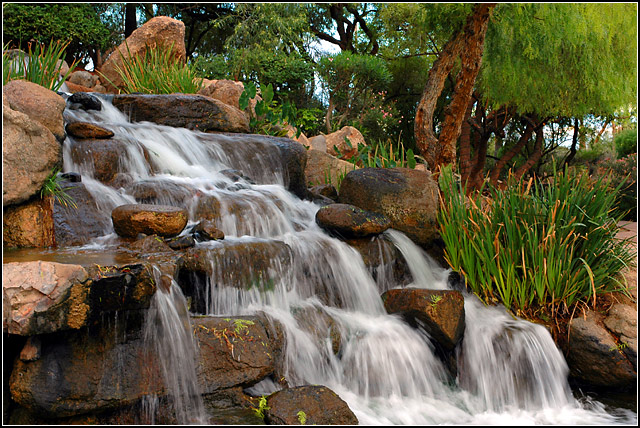} }
\subfigure[\scriptsize 5.71 (5.78)]{
\includegraphics*[scale=0.13]{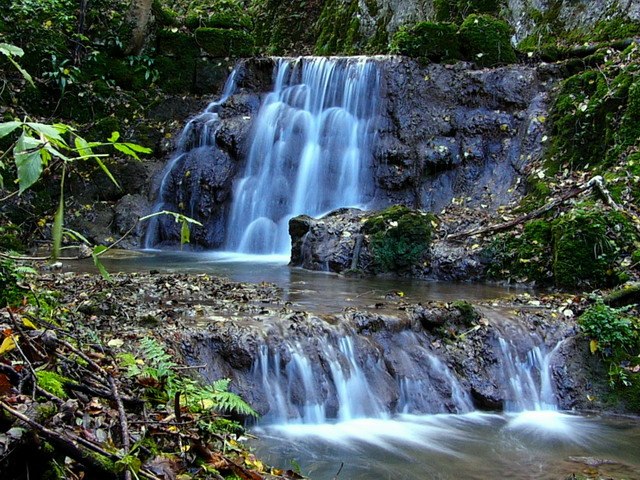} }
\subfigure[\scriptsize 5.61 (5.54)]{
\includegraphics*[scale=0.13]{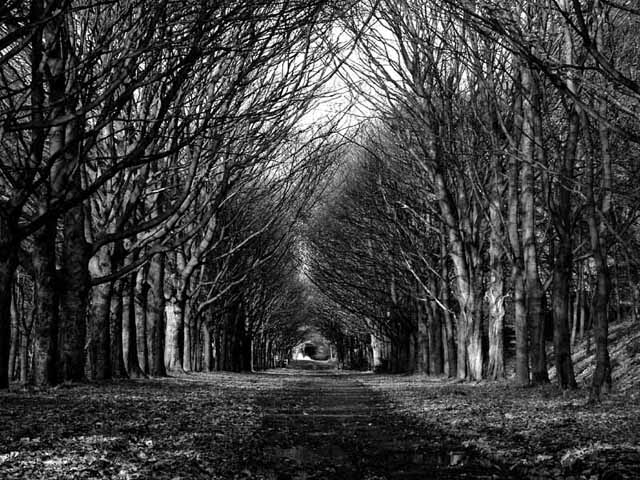} }
\subfigure[\scriptsize 5.28 (5.32)]{
\includegraphics*[scale=0.145]{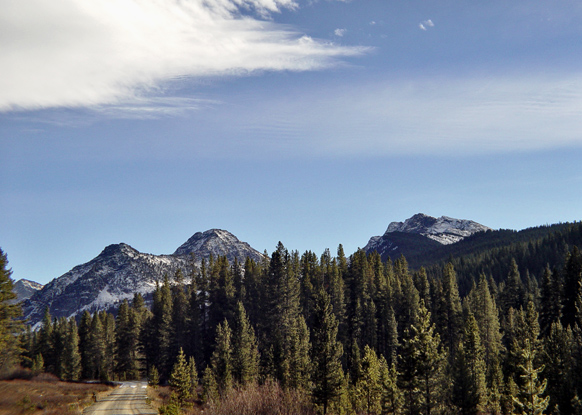} }
\subfigure[\scriptsize 5.11 (5.23)]{
\includegraphics*[scale=0.145]{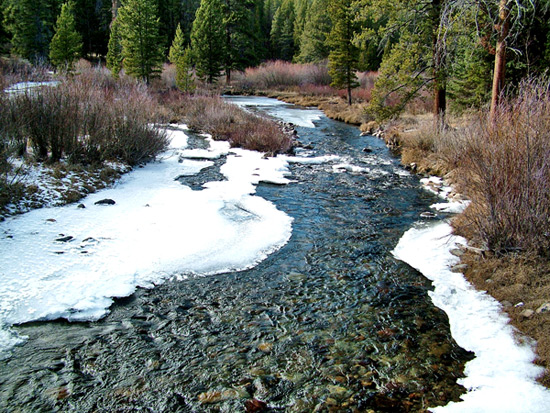} }
\subfigure[\scriptsize 5.03 (5.35)]{
\includegraphics*[scale=0.13]{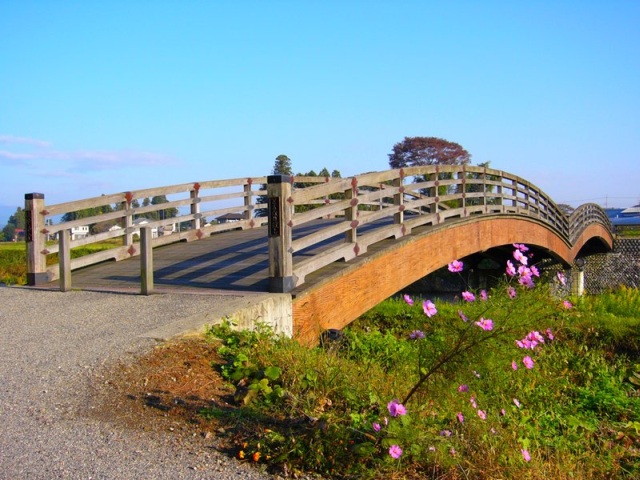} }
\subfigure[\scriptsize 4.90 (4.91)]{
\includegraphics*[scale=0.13]{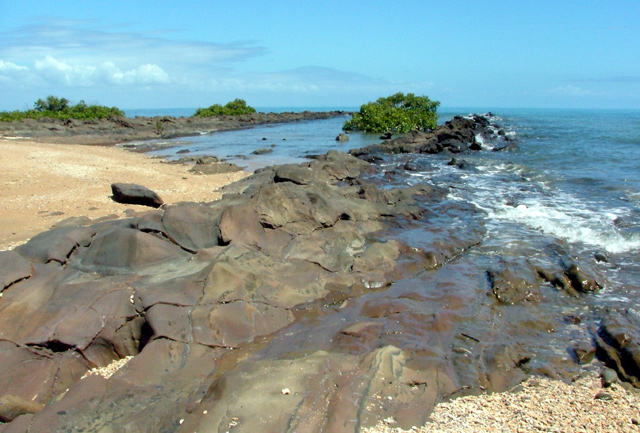} }
\subfigure[\scriptsize 4.83 (4.89)]{
\includegraphics*[scale=0.17]{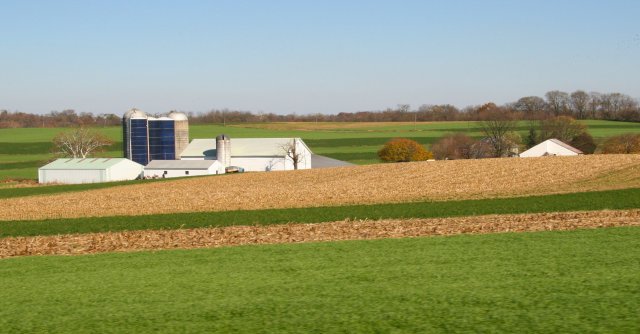} }
\subfigure[\scriptsize 4.77 (4.55)]{
\includegraphics*[scale=0.145]{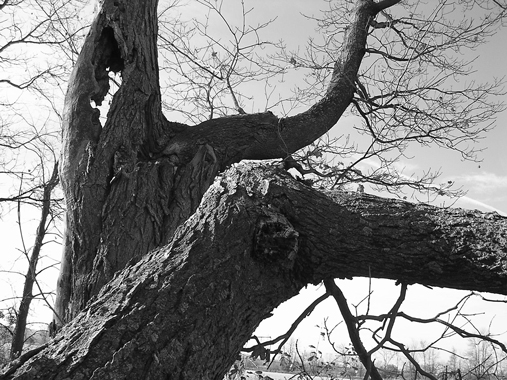} }
\subfigure[\scriptsize 4.48 (3.95)]{
\includegraphics*[scale=0.235]{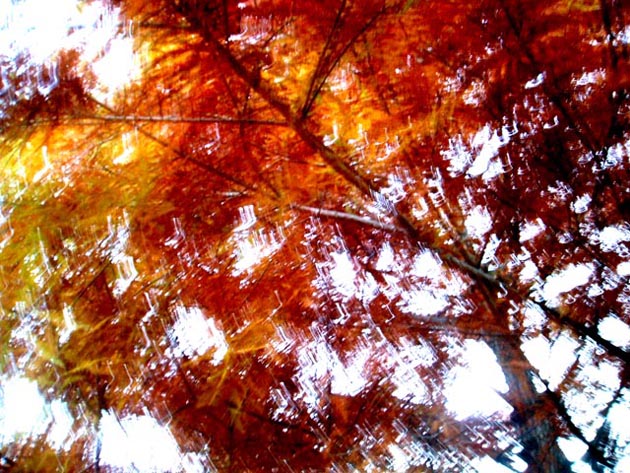} }
\subfigure[\scriptsize 3.55 (3.53)]{
\includegraphics*[scale=0.275]{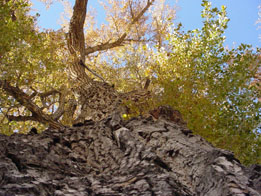} }
\end{center}
\vspace{-0 mm}
{\caption{Ranking some examples labelled with ``\textit{landscape}" tag from AVA dataset\cite{murray2012ava} using our proposed aesthetic assessment model NIMA(VGG16). Predicted (and ground truth) scores are shown below each image.\label{fig:ava_rank_landscape}}}
\vspace{-0 mm}
\end{figure*}

\begin{figure*}[!t]
\vspace{4 mm}
\begin{center}
\subfigure[\scriptsize 6.88 (7.40)]{
\includegraphics*[scale=0.23]{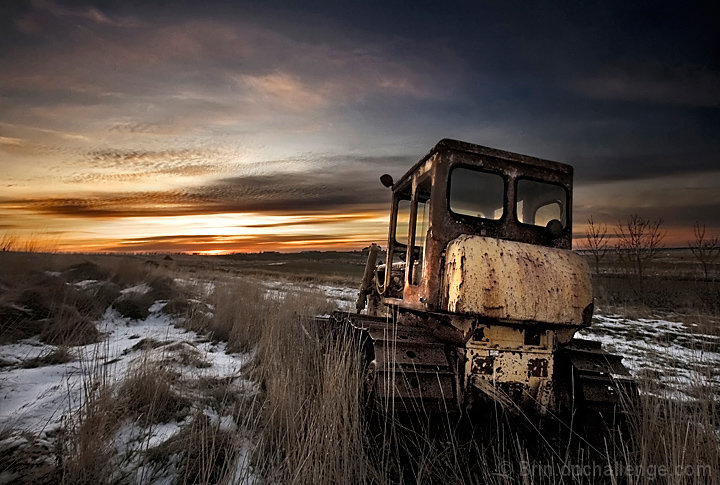} }
\subfigure[\scriptsize 6.63 (6.89)]{
\includegraphics*[scale=0.175]{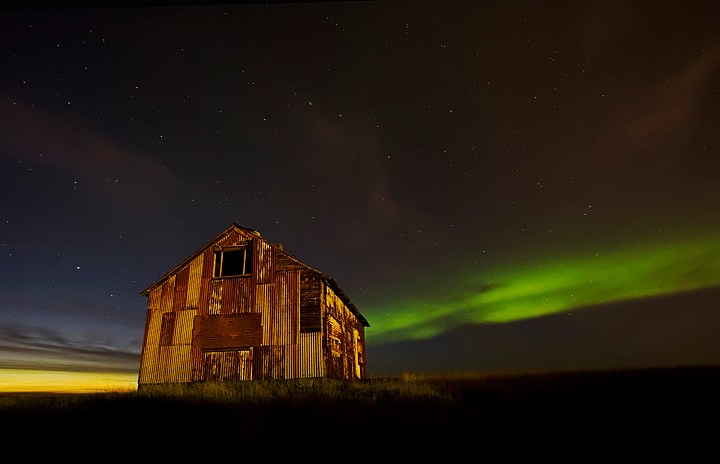} }
\subfigure[\scriptsize 6.29 (6.59)]{
\includegraphics*[scale=0.13]{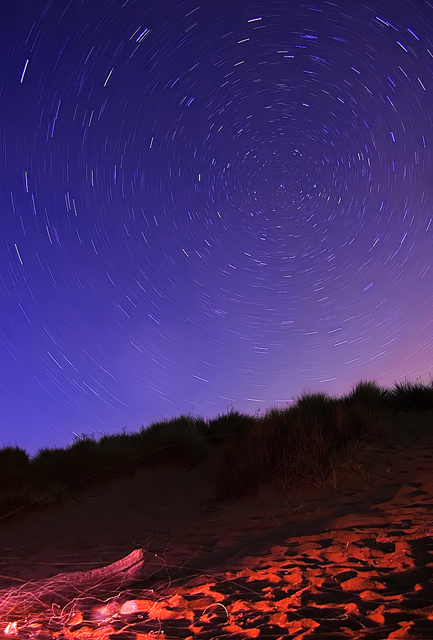} }
\subfigure[\scriptsize 5.86 (6.16)]{
\includegraphics*[scale=0.157]{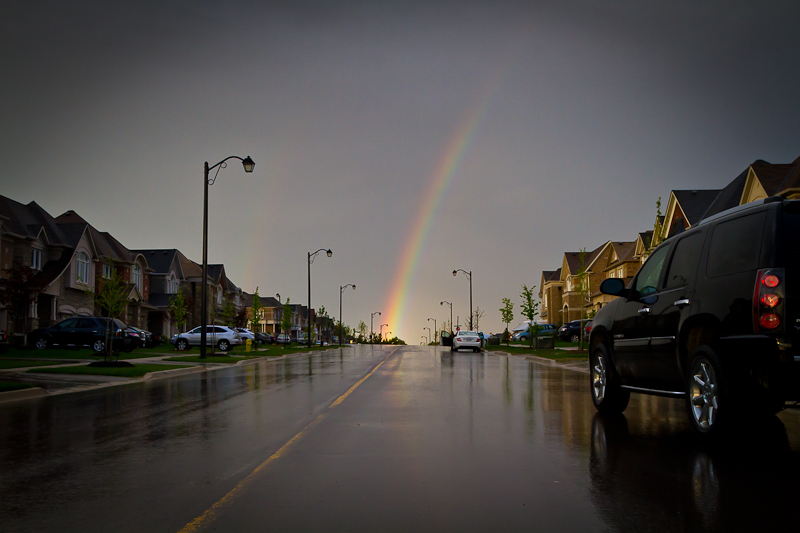} }
\subfigure[\scriptsize 5.77 (5.52)]{
\includegraphics*[scale=0.65]{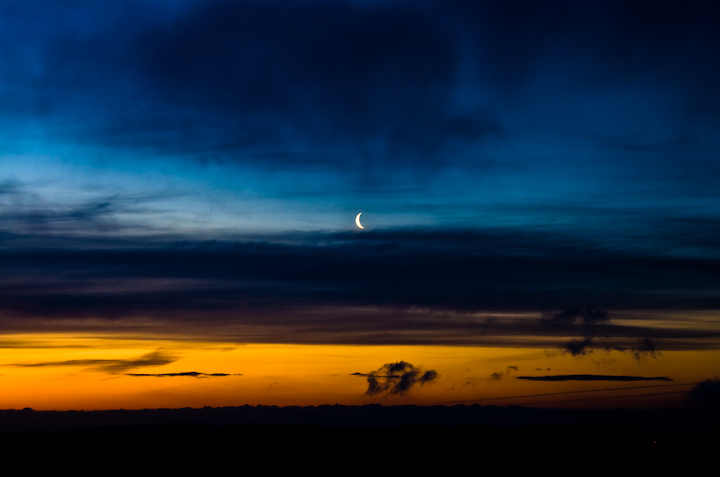} }
\subfigure[\scriptsize 5.51 (5.47)]{
\includegraphics*[scale=0.25]{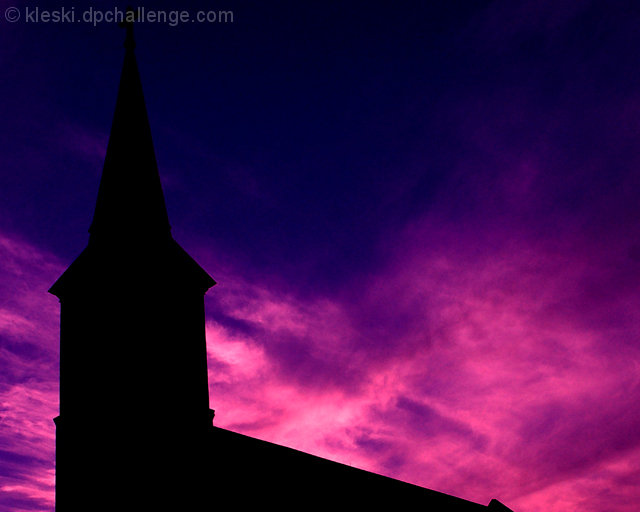} }
\subfigure[\scriptsize 5.46 (5.38)]{
\includegraphics*[scale=0.2]{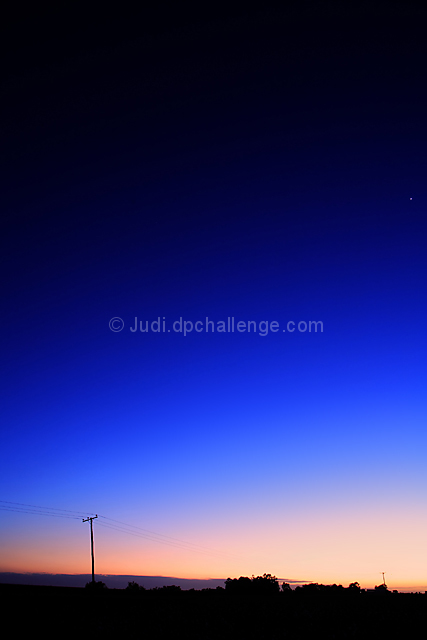} }
\subfigure[\scriptsize 5.24 (4.74)]{
\includegraphics*[scale=0.145]{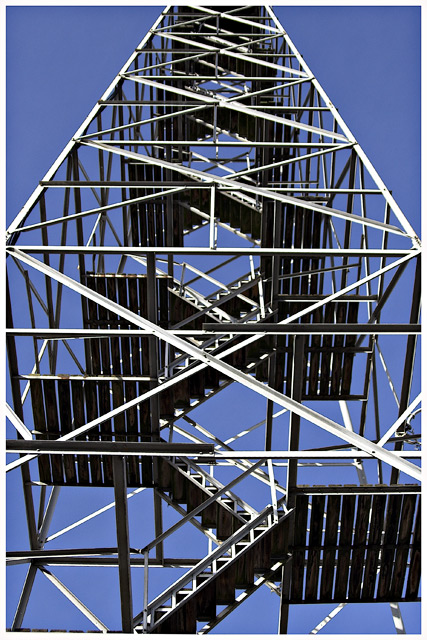} }
\subfigure[\scriptsize 4.96 (4.83) ]{
\includegraphics*[scale=0.74]{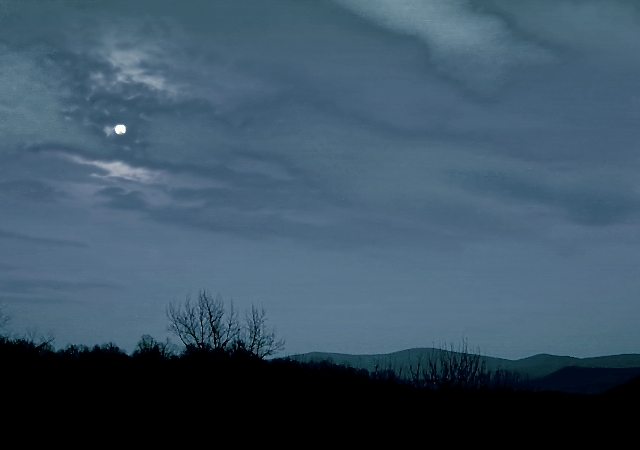} }
\subfigure[\scriptsize 4.90 (4.71) ]{
\includegraphics*[scale=0.64]{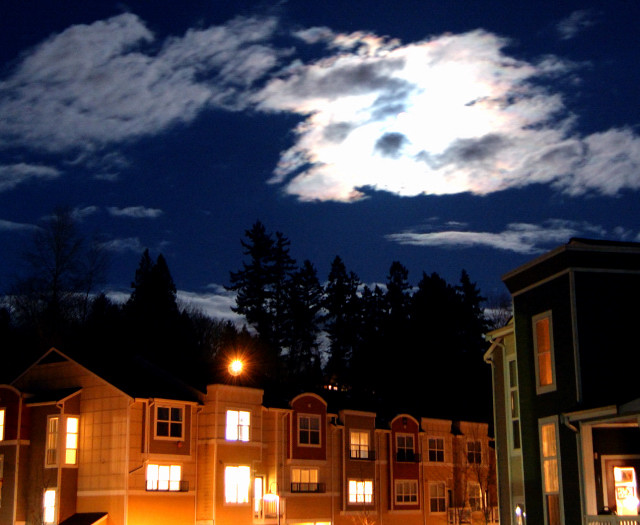} }
\subfigure[\scriptsize 4.60 (4.59) ]{
\includegraphics*[scale=0.195]{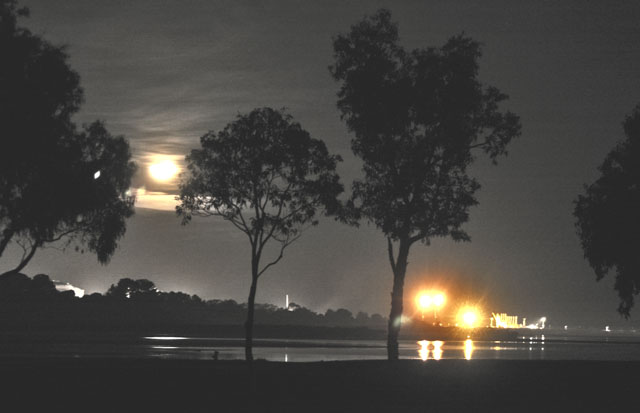} }
\subfigure[\scriptsize 4.53 (5.05) ]{
\includegraphics*[scale=0.135]{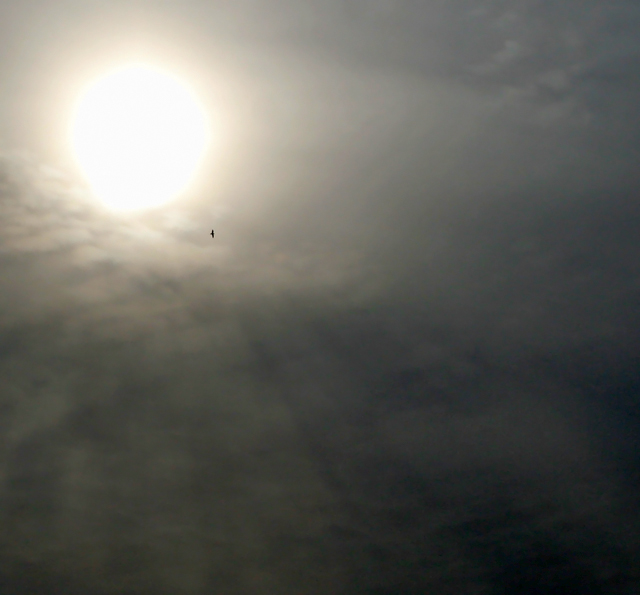} }
\end{center}
\vspace{-0 mm}
{\caption{Ranking some examples labelled with ``\textit{sky}" tag from AVA dataset\cite{murray2012ava} using our proposed aesthetic assessment model NIMA(Inception-v2). Predicted (and ground truth) scores are shown below each image.\label{fig:ava_rank_sky}}}
\vspace{4 mm}
\end{figure*}

Predicted mean scores can be used to rank photos, aesthetically. Some test photos from AVA dataset are ranked in Fig.~\ref{fig:ava_rank_landscape} and Fig.~\ref{fig:ava_rank_sky}. Predicted NIMA scores and ground truth AVA scores are shown below each image. Results in Fig.~\ref{fig:ava_rank_landscape} suggest that in addition to image content, other factors such as tone, contrast and composition of photos are important aesthetic qualities. Also, as shown in Fig.~\ref{fig:ava_rank_sky}, besides image semantics, framing and color palette are key qualities in these photos. These aesthetic attributes are closely predicted by our trained models on AVA.

\begin{figure*}[!t]
\vspace{4 mm}
\begin{center}
\subfigure[\scriptsize 5.31 (5.93)]{
\includegraphics*[scale=0.205]{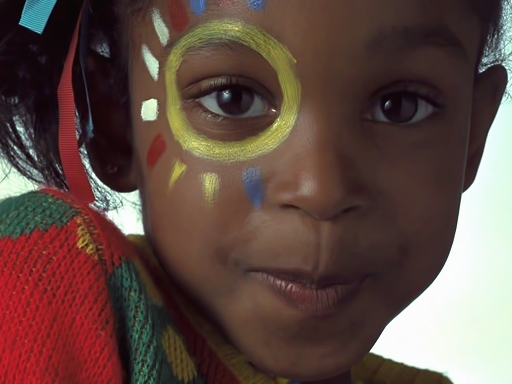} }
\subfigure[\scriptsize 4.35 (4.64)]{
\includegraphics*[scale=0.205]{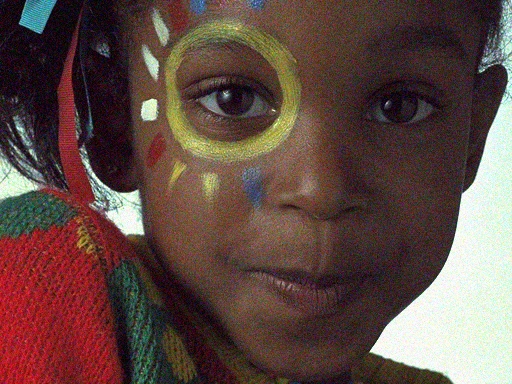} }
\subfigure[\scriptsize 4.00 (3.91)]{
\includegraphics*[scale=0.205]{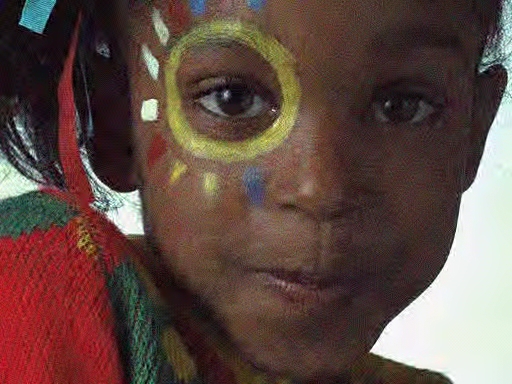} }
\subfigure[\scriptsize 3.56 (3.61)]{
\includegraphics*[scale=0.205]{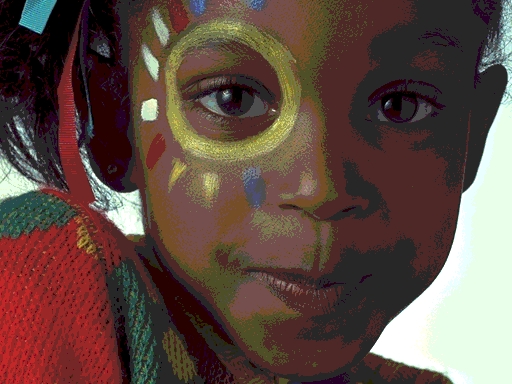} }
\subfigure[\scriptsize 3.05 (3.26)]{
\includegraphics*[scale=0.205]{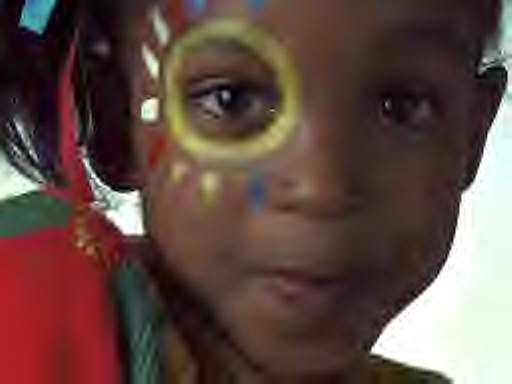} }
\subfigure[\scriptsize 2.87 (2.86)]{
\includegraphics*[scale=0.205]{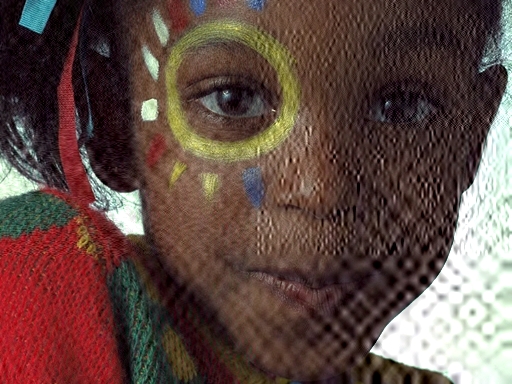} }
\subfigure[\scriptsize 2.33 (2.44)]{
\includegraphics*[scale=0.205]{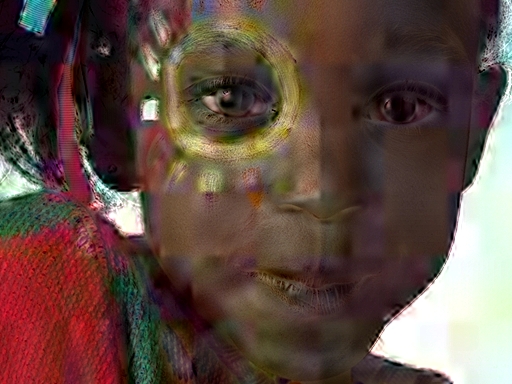} }
\subfigure[\scriptsize 1.67 (0.73)]{
\includegraphics*[scale=0.205]{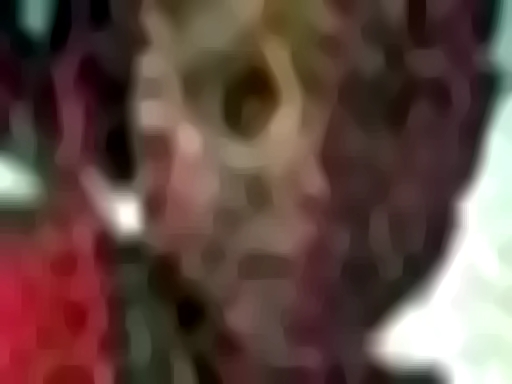} }
\end{center}
\vspace{-0 mm}
{\caption{Ranking some examples from TID2013 dataset\cite{ponomarenko2013color} using our proposed quality assessment model NIMA(VGG16). Predicted (and ground truth) scores are shown below each image.\label{fig:tid_rank}}}
\vspace{4 mm}
\end{figure*}

Predicted mean scores are used to qualitatively rank photos in Fig.~\ref{fig:tid_rank}. These images are part of our TID2013 test set, which contain various types and levels of distortions. Comparing ground truth and predicted scores indicates that our trained model on TID2013 accurately ranks the test images.

\begin{table}[!t]
\vspace{0 mm}
\begin{center}
\captionsetup{width=0.49\textwidth}
\caption{Performance of the proposed method with various architectures in predicting TID2013 quality ratings~\cite{ponomarenko2013color} compared to the state-of-the-art. LCC (linear correlation coefficient) and SRCC (Spearman's rank correlation coefficient) are computed between predicted and ground truth mean scores (column 2 and 3) and standard deviation of scores (column 4 and 5). EMD measures closeness of the predicted and ground truth rating distributions with $r=1$ in Eq. \ref{eqn:emd}. The LCC, and SROC values are in  $\pm0.005$, and $\pm0.004$ within $95\%$ confidence, respectively.}
\begin{tabular}{@{} *6l @{}}    \toprule
\emph{Model} & \emph{LCC} & \emph{SRCC} & \emph{LCC} & \emph{SRCC} & \emph{EMD} \\
  & \emph{(mean)} & \emph{(mean)} & \emph{(std.dev)} & \emph{(std.dev)}\\\midrule
 Kim et al.  \cite{kim2017deep} & 0.80 & 0.80 & -- & -- & -- \\
Moorthy et al. \cite{moorthy2011blind}    &  0.89  &  0.88 &  --  &  -- &  --   \\ 
Mittal et al. \cite{mittal2012no}    &  0.92  &  0.89  &  --  &  --  &  --  \\ 
Saad et al. \cite{saad2012blind}   &  0.91  &  0.88  &  --  &  --  &  --  \\ 
Kottayil et al. \cite{kottayil2016color}    &  0.89  &  0.88  &  --  &  --  &  --  \\ 
Xu et al. \cite{xu2016blind}    &  0.96  &  0.95  &  --  &  -- &  --   \\ 
Bianco et al. \cite{bianco2016use}   &  0.96  &  0.96  &  --  &  --  &  --  \\ \hdashline
NIMA(MobileNet)  & 0.782 & 0.698 &  0.209 &  0.181 &  0.105 \\
NIMA(VGG16)  & 0.941& 0.944 &  0.538 &  0.557 &  0.054 \\
NIMA(Inception-v2)  & 0.827 & 0.750 &  0.470 &  0.468 &  0.064 \\\bottomrule
 \hline
\end{tabular}
\label{tab:tid_comp}
\end{center}
\vspace{-5 mm}
\end{table}

\subsection{Image Enhancement}
\label{sec:enhancement}
\vspace{0 mm}

Quality and aesthetic scores can be used to perceptually tune image enhancement operators. In other words, maximizing NIMA score as a prior can increase the likelihood of enhancing perceptual quality of an image. Typically, parameters of enhancement operators such as image denoising and contrast enhancement are selected by extensive experiments under various photographic conditions. Perceptual tuning could be quite expensive and time consuming, especially when human opinion is required. In this section, our proposed models are used to tune a tone enhancement method~\cite{talebi2016fast}, and an image denoiser~\cite{wong2016turbo}. A more detailed treatment is presented in \cite{talebi2017learned}.

\begin{figure*}[!t]
\vspace{12 mm}
\begin{center}
\subfigure[\scriptsize Input (5.52)]{
\includegraphics*[scale=0.285]{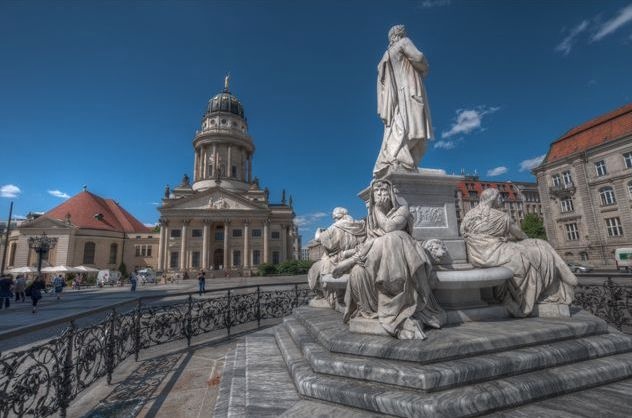} }
\subfigure[\scriptsize contrast compression (4.79)]{
\includegraphics*[scale=0.285]{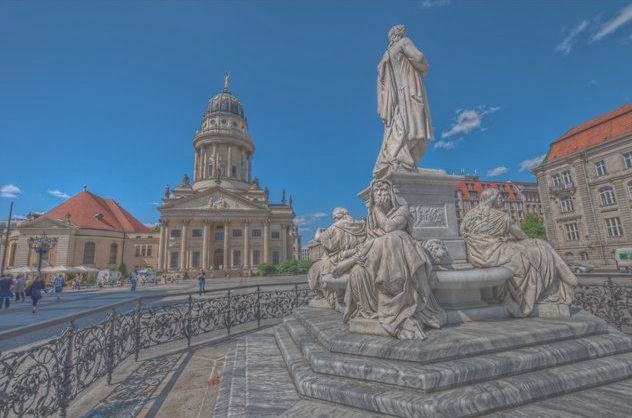} }
\subfigure[\scriptsize boosting details (5.73)]{
\includegraphics*[scale=0.285]{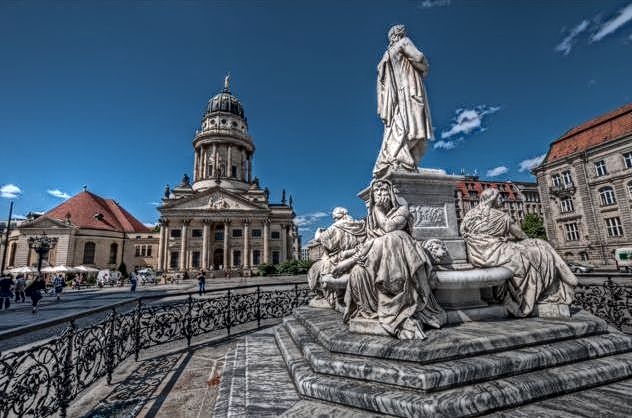} }
\subfigure[\scriptsize increasing brightness (5.52)]{
\includegraphics*[scale=0.285]{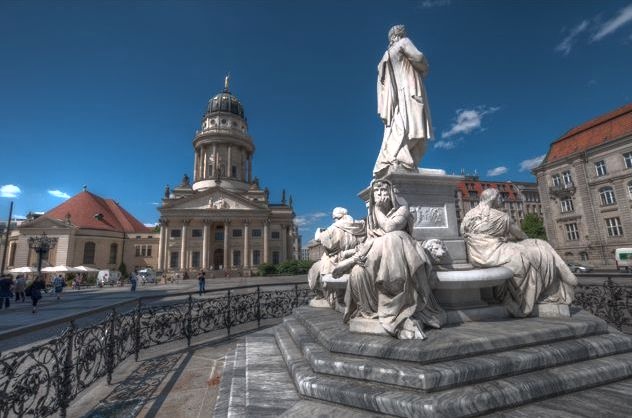} }
\subfigure[\scriptsize increasing shadows (5.95)]{
\includegraphics*[scale=0.285]{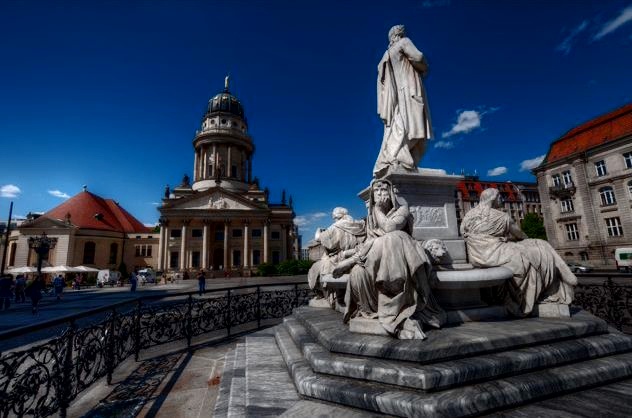} }
\end{center}
\vspace{-0 mm}
{\caption{Predicted aesthetic score (NIMA(VGG16)) for various parameter settings of multi-layer Laplacian technique~\cite{talebi2016fast}. Predicted aesthetic scores are shown below each image. \label{fig:ltm_results2}}}
\vspace{12 mm}
\end{figure*}

\begin{figure*}[!t]
\vspace{8 mm}
\begin{center}
\subfigure[\scriptsize Input (5.80)]{
\includegraphics*[scale=0.19]{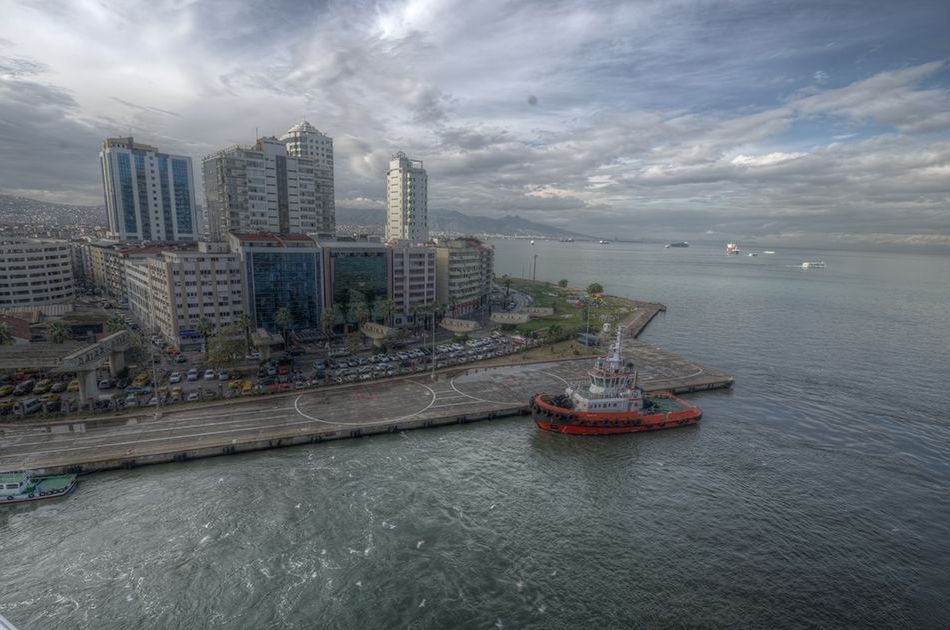} }
\subfigure[\scriptsize Enhanced (6.12)]{
\includegraphics*[scale=0.19]{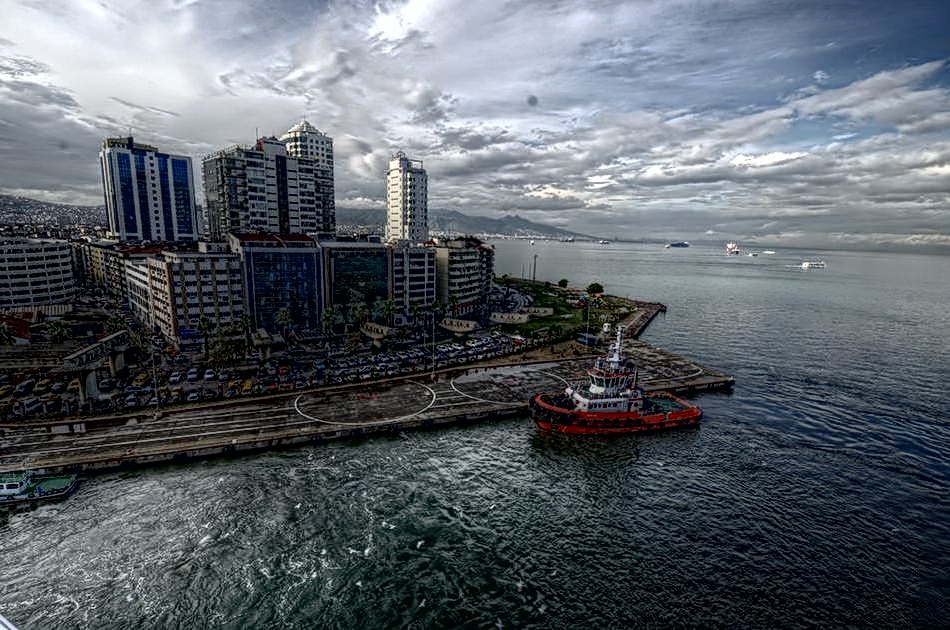} }
\subfigure[\scriptsize Input (5.52)]{
\includegraphics*[scale=0.285]{./figures/14_0_50_50.jpg} }
\subfigure[\scriptsize Enhanced (6.13)]{
\includegraphics*[scale=0.285]{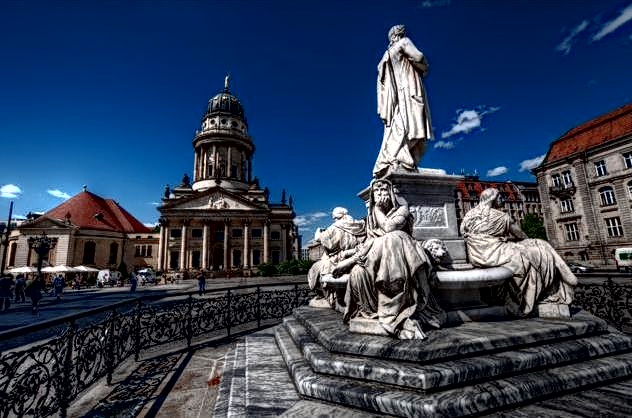} }
\subfigure[\scriptsize Input (4.87)]{
\includegraphics*[scale=0.46]{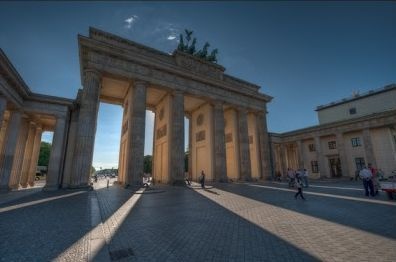} }
\subfigure[\scriptsize Enhanced (5.57)]{
\includegraphics*[scale=0.46]{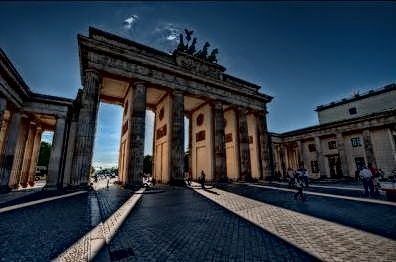} }
\subfigure[\scriptsize Input (5.59)]{
\includegraphics*[scale=0.29]{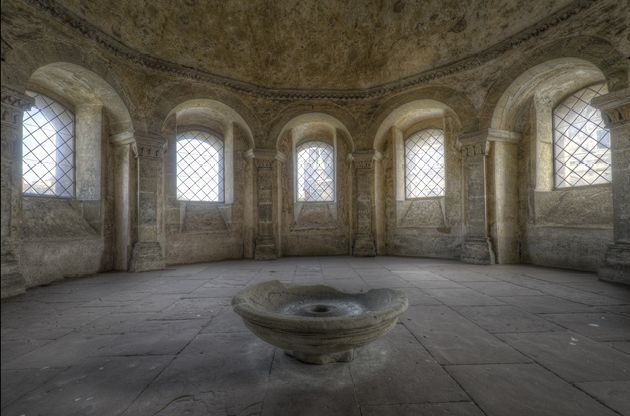} }
\subfigure[\scriptsize Enhanced (5.98)]{
\includegraphics*[scale=0.29]{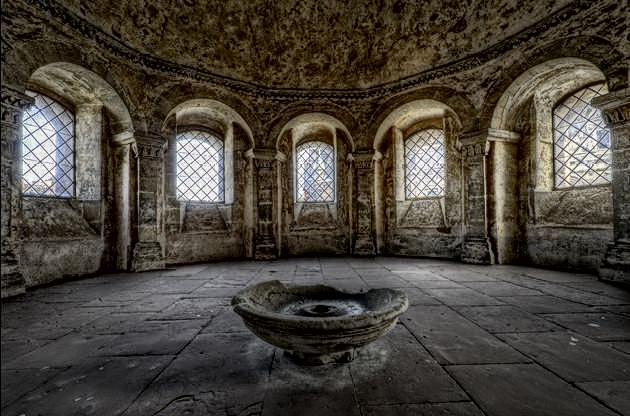} }
\end{center}
\vspace{0 mm}
{\caption{Tone enhancement by multi-layer Laplacian technique~\cite{talebi2016fast} along with our proposed aesthetic assessment model NIMA(VGG16). Predicted aesthetic scores are shown below each image. (Input photos are downloaded from \hyperref[www.farbspiel-photo.com]{www.farbspiel-photo.com}) \label{fig:ltm_results}}}
\vspace{8 mm}
\end{figure*}

The multi-layer Laplacian technique~\cite{talebi2016fast} enhances local and global contrast of images. Parameters of this method control the amount of detail, shadow, and brightness of an image. Fig.~\ref{fig:ltm_results2} shows a few examples of the multi-layer Laplacian with different sets of parameters. We observed that the predicted aesthetic ratings from training on the AVA dataset can be improved by contrast adjustments. Consequently, our model is able to guide the multi-layer Laplacian filter to find aesthetically near-optimal settings of its parameters. Examples of this type of image editing are represented in Fig.~\ref{fig:ltm_results}, where a combination of detail, shadow and brightness change is applied on each image. In each example, 6 levels of detail boost, 11 levels of shadow change, and 11 levels of brightness change account for a total of 726 variations. The aesthetic assessment model tends to prefer high contrast images with boosted details. This is consistent with the ground truth results from AVA illustrated in Fig.~\ref{fig:ava_rank_landscape}.

\begin{figure*}[!t]
\vspace{-5 mm}
\begin{center}
\subfigure[\scriptsize Noisy Input]{
\includegraphics*[scale=0.28]{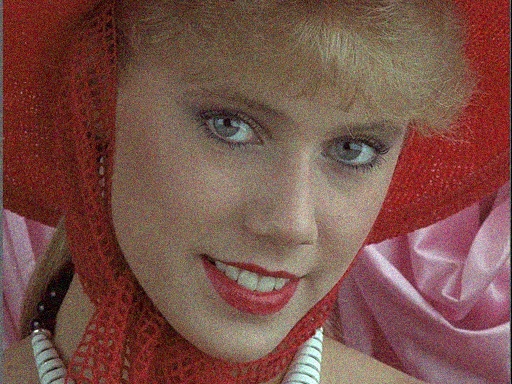} }
\subfigure[\scriptsize Optimized (denoising parameter=3.75)]{
\includegraphics*[scale=0.28]{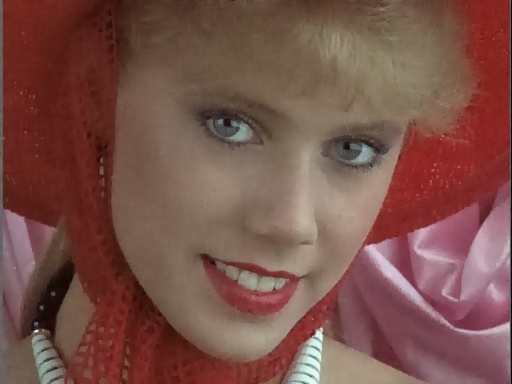} }
\subfigure[\scriptsize Quality score vs. denoising parameter]{
\includegraphics*[viewport=30 10 520 400, scale=0.285]{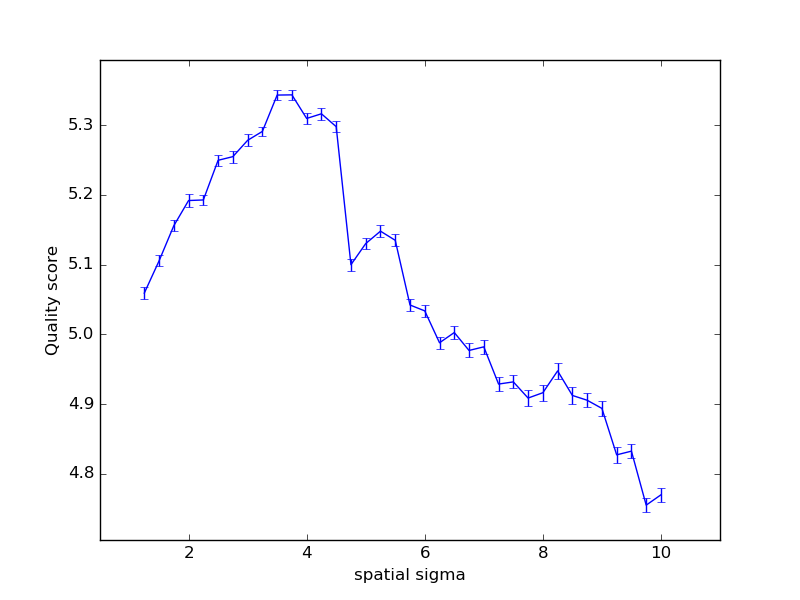} }
\subfigure[\scriptsize Noisy Input]{
\includegraphics*[scale=0.28]{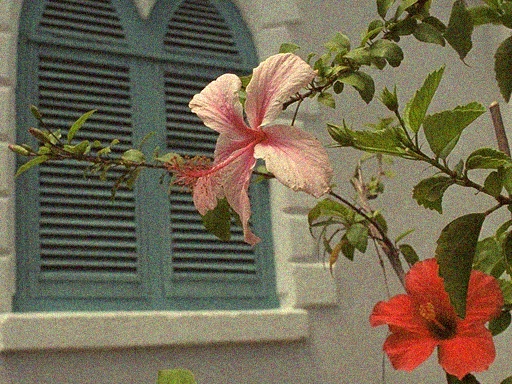} }
\subfigure[\scriptsize Optimized (denoising parameter=2.25)]{
\includegraphics*[scale=0.28]{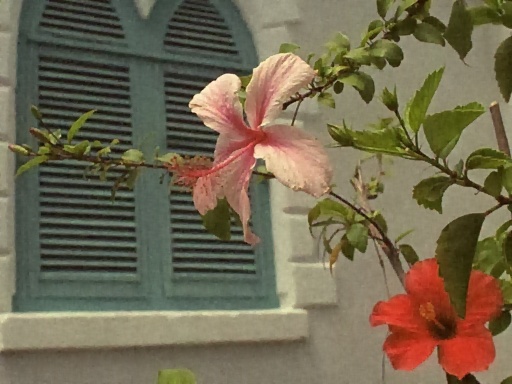} }
\subfigure[\scriptsize Quality score vs. denoising parameter]{
\includegraphics*[viewport=30 10 520 400, scale=0.285]{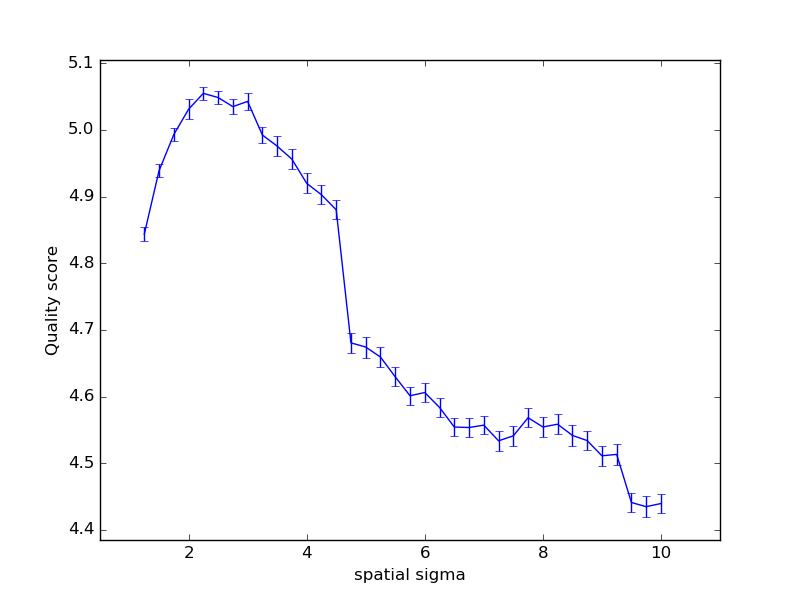} }
\subfigure[\scriptsize Noisy Input]{
\includegraphics*[scale=0.28]{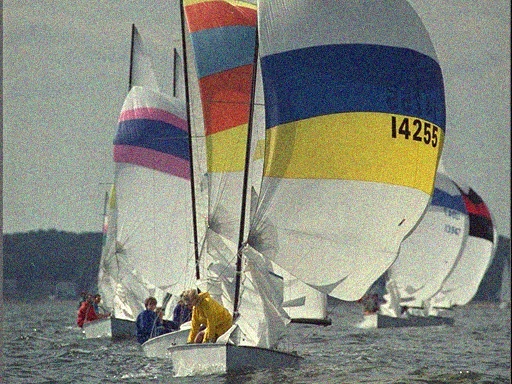} }
\subfigure[\scriptsize Optimized (denoising parameter=4.50)]{
\includegraphics*[scale=0.28]{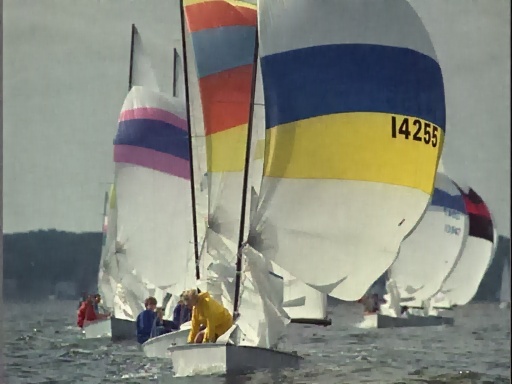} }
\subfigure[\scriptsize Quality score vs. denoising parameter]{
\includegraphics*[viewport=30 10 520 400, scale=0.285]{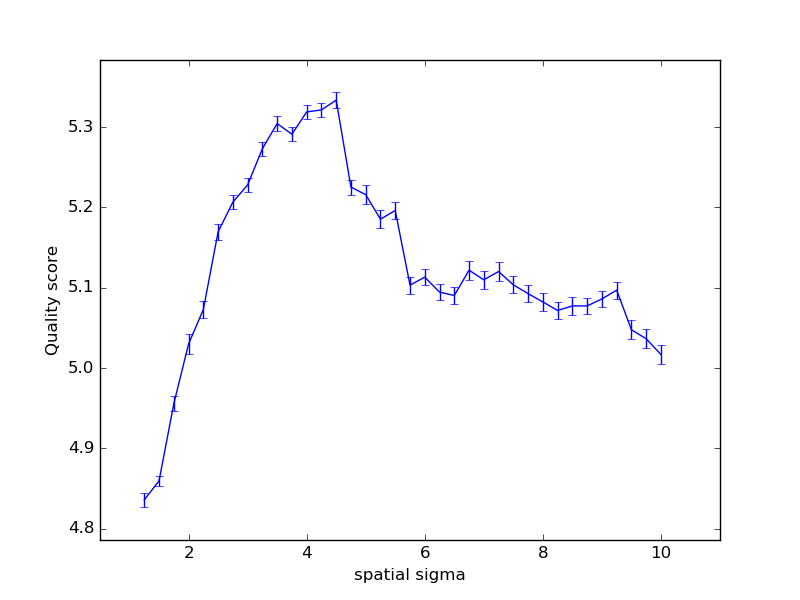} }
\subfigure[\scriptsize Noisy Input]{
\includegraphics*[scale=0.28]{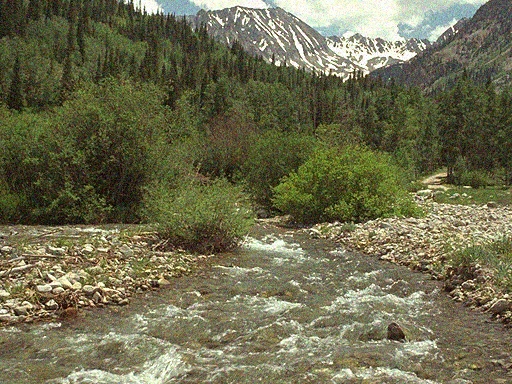} }
\subfigure[\scriptsize Optimized (denoising parameter=1.25)]{
\includegraphics*[scale=0.28]{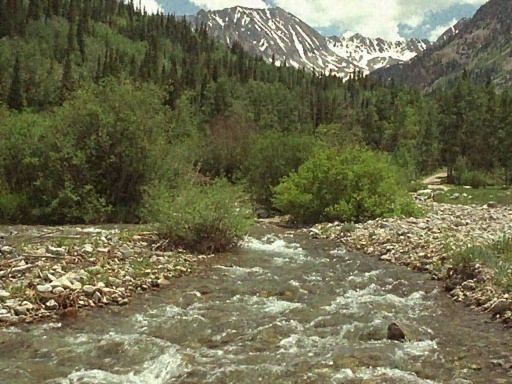} }
\subfigure[\scriptsize Quality score vs. denoising parameter]{
\includegraphics*[viewport=30 10 520 400, scale=0.285]{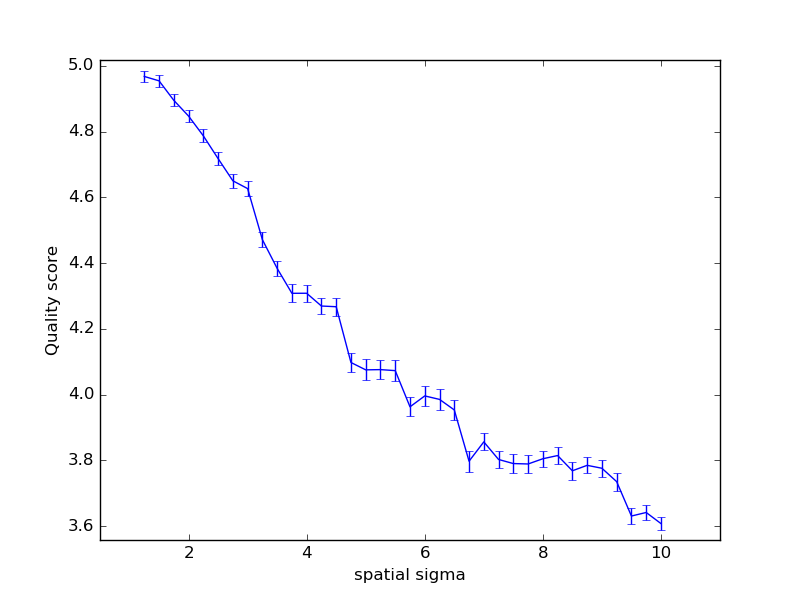} }
\end{center}
\vspace{-5 mm}
{\caption{Tuning spatial parameter of Turbo denoising~\cite{wong2016turbo} by using our proposed quality assessment model NIMA(VGG16). Standard deviation of the additive white Gaussian noise is set as 30. Denoised results are shown for maximum quality score. \label{fig:denoiser_results}}}
\vspace{-0 mm}
\end{figure*}

\begin{figure*}[!t]
\vspace{-0 mm}
\begin{center}
\subfigure[\scriptsize denoising parameter=1.25, score=5.06]{
\includegraphics*[viewport=30 10 320 400, scale=0.5]{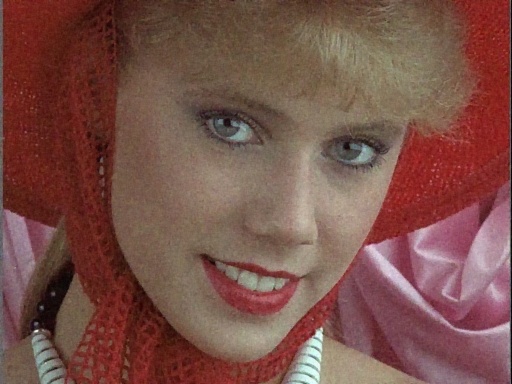} }
\subfigure[\scriptsize denoising parameter=3.0, score=5.15]{
\includegraphics*[viewport=30 10 320 400, scale=0.5]{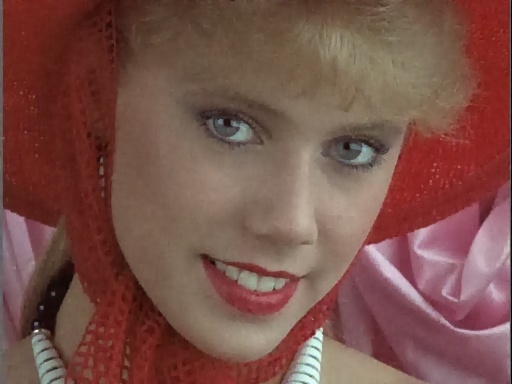} }
\subfigure[\scriptsize denoising parameter=9.75, score=4.76]{
\includegraphics*[viewport=30 10 320 400, scale=0.5]{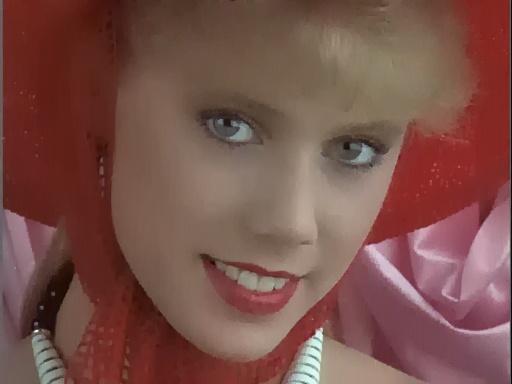} }
\subfigure[\scriptsize denoising parameter=1.25, score=4.97]{
\includegraphics*[viewport=90 70 320 400, scale=0.62]{./figures/I13_sigma_spatial=1250000.jpg} }
\subfigure[\scriptsize denoising parameter=3.0, score=4.62]{
\includegraphics*[viewport=90 70 320 400, scale=0.62]{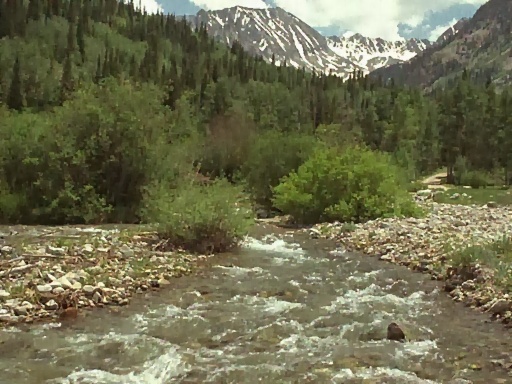} }
\subfigure[\scriptsize denoising parameter=9.75, score=3.64]{
\includegraphics*[viewport=90 70 320 400, scale=0.62]{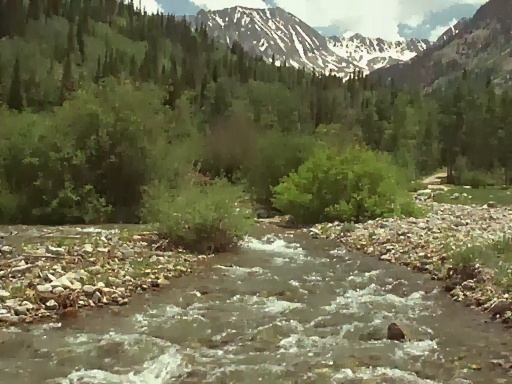} }
\end{center}
\vspace{-3 mm}
{\caption{Effect of Turbo denoising~\cite{wong2016turbo} on our predicted quality scores. Input noisy images are shown in Fig.~\ref{fig:denoiser_results}. \label{fig:denoiser_results2}}}
\vspace{-0 mm}
\end{figure*}

Turbo denoising~\cite{wong2016turbo} is a technique which uses the domain transform~\cite{gastal2011domain} as its core filter. Performance of Turbo denoising depends on spatial and range smoothing parameters, and consequently, proper tuning of these parameters can effectively boost performance of the denoiser. We observed that varying the spatial smoothing parameter makes the most significant perceptual difference, and as a result, we use our quality assessment model trained on TID2013 dataset to tune this denoiser. Application of our no-reference quality metric as a prior in image denoising is similar to the work of Zhu et al. ~\cite{zhu2009no, zhu2010automatic}. Our results are shown in Fig.~\ref{fig:denoiser_results}. Additive white Gaussian noise with standard deviation 30 is added to the clean image, and Turbo denoising with various spatial parameters is used to denoise the noisy image. To reduce the score deviation, 50 random crops are extracted from denoised image. These scores are averaged to obtain the plots illustrated in Fig.~\ref{fig:denoiser_results}. As can be seen, although the same amount of noise is added to each image, maximum quality scores correspond to different denoising parameters in each example. For relatively smooth images such as (a) and (g), optimal spatial parameter of Turbo denoising is higher (which implies stronger smoothing) than the textured image in (j). This is probably due to the relatively high signal-to-noise ratio of (j). In other words, the quality assessment model tends to respect textures and avoid over-smoothing of details. Effect of the denoising parameter can be visually inspected in Fig.~\ref{fig:denoiser_results2}. While the denoised result in Fig.~\ref{fig:denoiser_results2} (a) is under-smoothed, (c), (e) and (f) show undesirable over-smoothing effects. The predicted quality scores validate this perceptual observation.

\begin{table}[!t]
\begin{center}
\captionsetup{width=0.48\textwidth}
\caption{LCC (linear correlation coefficient) of NIMA(Inception-v2) model for training and testing on various datasets.}
\begin{tabular}{c|cccc}    \toprule
 & & Test Dataset & \\ \cline{2-4}
Train Dataset  & LIVE~\cite{ghadiyaram2016massive} & TID2013~\cite{ponomarenko2013color}  & AVA~\cite{murray2012ava} & Average \\\midrule
LIVE~\cite{ghadiyaram2016massive} & 0.698 & 0.537 &   0.238 & 0.491\\ 
TID2013~\cite{ponomarenko2013color}  & 0.178 & 0.827 &   0.101 &  0.369\\
AVA~\cite{murray2012ava} & 0.552 & 0.514 & 0.636 & \textbf{0.567}\\\bottomrule
 \hline
\end{tabular}
\label{tab:cross_validation_lcc}
\end{center}
\end{table}

\begin{table}[!t]
\begin{center}
\captionsetup{width=0.48\textwidth}
\caption{SRCC (Spearman's rank correlation coefficient) of NIMA(Inception-v2) model for training and testing on various datasets.}
\begin{tabular}{c|cccc}    \toprule
 & & Test Dataset & \\ \cline{2-4}
Train Dataset  & LIVE~\cite{ghadiyaram2016massive} & TID2013~\cite{ponomarenko2013color} & AVA~\cite{murray2012ava} & Average \\\midrule
LIVE~\cite{ghadiyaram2016massive} & 0.637 & 0.327 &   0.200 & 0.388\\ 
TID2013~\cite{ponomarenko2013color} & 0.155 & 0.750 &   0.087 &  0.331\\
AVA~\cite{murray2012ava} & 0.543 & 0.432 & 0.612 & \textbf{0.529}\\\bottomrule
 \hline
\end{tabular}
\label{tab:cross_validation_srcc}
\end{center}
\end{table}

\subsection{Computational Costs}
\label{sec:comput_cost}
\vspace{0 mm}

Computational complexity of NIMA models are compared in Table~\ref{tab:time_comp}. Our inference TensorFlow implementation is tested on an Intel Xeon CPU @ 3.5 GHz with 32 GB memory and 12 cores, and NVIDIA Quadro K620 GPU. Timings of one pass of NIMA models on an image of size $224\times224\times3$ are reported in Table~\ref{tab:time_comp}. Evidently, NIMA(MobileNet) is significantly lighter and faster than other models. This comes at the expense of a slight performance drop (shown in Table~\ref{tab:ava_comp} and Table~\ref{tab:tid_comp}).

\section{Conclusion}
\label{sec:conclusion}
\vspace{0 mm}

In this work we introduced a CNN-based image assessment method, which can be trained on both aesthetic and pixel-level quality datasets. Our models effectively predict the distribution of quality ratings, rather than just the mean scores. This leads to a more accurate quality prediction with higher correlation to the ground truth ratings. We trained two models for high level aesthetics and low level technical qualities, and utilized them to steer parameters of a few image enhancement operators. Our experiments suggest that these models are capable of guiding denoising and tone enhancement to produce perceptually superior results. 

As part of our future work, we will exploit the trained models on other image enhancement applications. Our current experimental setup requires the enhancement operator to be evaluated multiple times. This limits real-time application of the proposed method. One might argue that in case of an enhancement operator with well-defined derivatives, using NIMA as the loss function is a more efficient approach.

\section*{Acknowledgment}
We would like to thank Dr. Pascal Getreuer for valuable discussions and helpful advice on approximation of score distributions.

\begin{table*}[!t]
\begin{center}
\captionsetup{width=0.5\textwidth}
\caption{Comparison of the proposed quality assessment technique with various CNN architectures. Average timings are reported in ms for Xeon Intel CPU @ 3.5 GHz, and NVIDIA Quadro K620 GPU. Timings are reported for applying NIMA models on images of size $224\times224\times3$.}
\begin{tabular}{@{} *5l @{}}    \toprule
 & \emph{Million} & \emph{Billion} & \emph{CPU} & \emph{GPU} \\
\emph{Model}  & \emph{Parameters} & \emph{Flops} & \emph{Timing (ms)}  & \emph{Timing (ms)} \\\midrule
NIMA(MobileNet)  & 3.22 & 1.29 &  30.45 &  20.23 \\
NIMA(Inception-v2)  & 10.16 & 4.37 &  70.49 &  39.11 \\
NIMA(VGG16)  & 134.30 & 31.62 &  150.34 &  85.76 \\\bottomrule
 \hline
\end{tabular}
\label{tab:time_comp}
\end{center}
\end{table*}

\bibliographystyle{IEEEtran}   
\bibliography{refs}

\begin{thebibliography}{10}
\providecommand{\url}[1]{#1}
\csname url@samestyle\endcsname
\providecommand{\newblock}{\relax}
\providecommand{\bibinfo}[2]{#2}
\providecommand{\BIBentrySTDinterwordspacing}{\spaceskip=0pt\relax}
\providecommand{\BIBentryALTinterwordstretchfactor}{4}
\providecommand{\BIBentryALTinterwordspacing}{\spaceskip=\fontdimen2\font plus
\BIBentryALTinterwordstretchfactor\fontdimen3\font minus
  \fontdimen4\font\relax}
\providecommand{\BIBforeignlanguage}[2]{{%
\expandafter\ifx\csname l@#1\endcsname\relax
\typeout{** WARNING: IEEEtran.bst: No hyphenation pattern has been}%
\typeout{** loaded for the language `#1'. Using the pattern for}%
\typeout{** the default language instead.}%
\else
\language=\csname l@#1\endcsname
\fi
#2}}
\providecommand{\BIBdecl}{\relax}
\BIBdecl

\bibitem{murray2012ava}
N.~Murray, L.~Marchesotti, and F.~Perronnin, ``{AVA}: A large-scale database
  for aesthetic visual analysis,'' in \emph{Computer Vision and Pattern
  Recognition (CVPR), 2012 IEEE Conference on}.\hskip 1em plus 0.5em minus
  0.4em\relax IEEE, 2012, pp. 2408--2415.

\bibitem{ponomarenko2013color}
N.~Ponomarenko, O.~Ieremeiev, V.~Lukin, K.~Egiazarian, L.~Jin, J.~Astola,
  B.~Vozel, K.~Chehdi, M.~Carli, F.~Battisti \emph{et~al.}, ``Color image
  database {TID2013}: Peculiarities and preliminary results,'' in \emph{Visual
  Information Processing (EUVIP), 2013 4th European Workshop on}.\hskip 1em
  plus 0.5em minus 0.4em\relax IEEE, 2013, pp. 106--111.

\bibitem{wang2004image}
Z.~Wang, A.~C. Bovik, H.~R. Sheikh, and E.~P. Simoncelli, ``Image quality
  assessment: from error visibility to structural similarity,'' \emph{IEEE
  Transactions on Image Processing}, vol.~13, no.~4, pp. 600--612, 2004.

\bibitem{xue2013learning}
W.~Xue, L.~Zhang, and X.~Mou, ``Learning without human scores for blind image
  quality assessment,'' in \emph{Proceedings of the IEEE Conference on Computer
  Vision and Pattern Recognition}, 2013, pp. 995--1002.

\bibitem{kang2014convolutional}
L.~Kang, P.~Ye, Y.~Li, and D.~Doermann, ``Convolutional neural networks for
  no-reference image quality assessment,'' in \emph{Proceedings of the IEEE
  Conference on Computer Vision and Pattern Recognition}, 2014, pp. 1733--1740.

\bibitem{bosse2016deep}
S.~Bosse, D.~Maniry, T.~Wiegand, and W.~Samek, ``A deep neural network for
  image quality assessment,'' in \emph{Image Processing (ICIP), 2016 IEEE
  International Conference on}.\hskip 1em plus 0.5em minus 0.4em\relax IEEE,
  2016, pp. 3773--3777.

\bibitem{bianco2016use}
S.~Bianco, L.~Celona, P.~Napoletano, and R.~Schettini, ``On the use of deep
  learning for blind image quality assessment,'' \emph{arXiv preprint
  arXiv:1602.05531}, 2016.

\bibitem{lu2015deep}
X.~Lu, Z.~Lin, X.~Shen, R.~Mech, and J.~Z. Wang, ``Deep multi-patch aggregation
  network for image style, aesthetics, and quality estimation,'' in
  \emph{Proceedings of the IEEE International Conference on Computer Vision},
  2015, pp. 990--998.

\bibitem{kao2015visual}
Y.~Kao, C.~Wang, and K.~Huang, ``Visual aesthetic quality assessment with a
  regression model,'' in \emph{Image Processing (ICIP), 2015 IEEE International
  Conference on}.\hskip 1em plus 0.5em minus 0.4em\relax IEEE, 2015, pp.
  1583--1587.

\bibitem{mai2016composition}
L.~Mai, H.~Jin, and F.~Liu, ``Composition-preserving deep photo aesthetics
  assessment,'' in \emph{Proceedings of the IEEE Conference on Computer Vision
  and Pattern Recognition}, 2016, pp. 497--506.

\bibitem{jin2016image}
B.~Jin, M.~V.~O. Segovia, and S.~S{\"u}sstrunk, ``Image aesthetic predictors
  based on weighted {CNN}s,'' in \emph{Image Processing (ICIP), 2016 IEEE
  International Conference on}.\hskip 1em plus 0.5em minus 0.4em\relax IEEE,
  2016, pp. 2291--2295.

\bibitem{sheikh2005live}
H.~R. Sheikh, Z.~Wang, L.~Cormack, and A.~C. Bovik, ``Live image quality
  assessment database release 2,'' 2005.

\bibitem{larson2010most}
E.~C. Larson and D.~M. Chandler, ``Most apparent distortion: full-reference
  image quality assessment and the role of strategy,'' \emph{Journal of
  Electronic Imaging}, vol.~19, no.~1, pp. 011\,006--011\,006, 2010.

\bibitem{kong2016photo}
S.~Kong, X.~Shen, Z.~Lin, R.~Mech, and C.~Fowlkes, ``Photo aesthetics ranking
  network with attributes and content adaptation,'' in \emph{European
  Conference on Computer Vision}.\hskip 1em plus 0.5em minus 0.4em\relax
  Springer, 2016, pp. 662--679.

\bibitem{krizhevsky2012imagenet}
A.~Krizhevsky, I.~Sutskever, and G.~E. Hinton, ``Image{N}et classification with
  deep convolutional neural networks,'' in \emph{Advances in neural information
  processing systems}, 2012, pp. 1097--1105.

\bibitem{kim2017deep}
J.~Kim, H.~Zeng, D.~Ghadiyaram, S.~Lee, L.~Zhang, and A.~C. Bovik, ``Deep
  convolutional neural models for picture-quality prediction: Challenges and
  solutions to data-driven image quality assessment,'' \emph{IEEE Signal
  Processing Magazine}, vol.~34, no.~6, pp. 130--141, 2017.

\bibitem{lu2015rating}
X.~Lu, Z.~Lin, H.~Jin, J.~Yang, and J.~Z. Wang, ``Rating image aesthetics using
  deep learning,'' \emph{IEEE Transactions on Multimedia}, vol.~17, no.~11, pp.
  2021--2034, 2015.

\bibitem{simonyan2014very}
K.~Simonyan and A.~Zisserman, ``Very deep convolutional networks for
  large-scale image recognition,'' \emph{arXiv preprint arXiv:1409.1556}, 2014.

\bibitem{zeng2017probabilistic}
H.~Zeng, L.~Zhang, and A.~C. Bovik, ``A probabilistic quality representation
  approach to deep blind image quality prediction,'' \emph{arXiv preprint
  arXiv:1708.08190}, 2017.

\bibitem{ma2017lamp}
S.~Ma, J.~Liu, and C.~W. Chen, ``A-lamp: Adaptive layout-aware multi-patch deep
  convolutional neural network for photo aesthetic assessment,'' in
  \emph{Computer Vision and Pattern Recognition (CVPR), 2017 IEEE Conference
  on}.\hskip 1em plus 0.5em minus 0.4em\relax IEEE, 2017.

\bibitem{hou2016squared}
L.~Hou, C.-P. Yu, and D.~Samaras, ``Squared earth mover's distance-based loss
  for training deep neural networks,'' \emph{arXiv preprint arXiv:1611.05916},
  2016.

\bibitem{zhang2018unreasonable}
R.~Zhang, P.~Isola, A.~A. Efros, E.~Shechtman, and O.~Wang, ``The unreasonable
  effectiveness of deep features as a perceptual metric,'' \emph{IEEE
  conference on Computer Vision and Pattern Recognition (CVPR)}, 2018.

\bibitem{talebi2017learned}
H.~Talebi and P.~Milanfar, ``Learned perceptual image enhancement,'' \emph{IEEE
  International Conference on Computational Photography (ICCP)}, May 2018.

\bibitem{gu2016analysis}
K.~Gu, G.~Zhai, W.~Lin, and M.~Liu, ``The analysis of image contrast: From
  quality assessment to automatic enhancement,'' \emph{IEEE transactions on
  cybernetics}, vol.~46, no.~1, pp. 284--297, 2016.

\bibitem{kodak_dataset}
Kodak, ``Kodak lossless true color image suite,''
  \url{http://r0k.us/graphics/kodak/}.

\bibitem{ghadiyaram2016massive}
D.~Ghadiyaram and A.~C. Bovik, ``Massive online crowdsourced study of
  subjective and objective picture quality,'' \emph{IEEE Transactions on Image
  Processing}, vol.~25, no.~1, pp. 372--387, 2016.

\bibitem{cover2012elements}
T.~M. Cover and J.~A. Thomas, \emph{Elements of information theory}.\hskip 1em
  plus 0.5em minus 0.4em\relax John Wiley \& Sons, 2012.

\bibitem{szegedy2016rethinking}
C.~Szegedy, V.~Vanhoucke, S.~Ioffe, J.~Shlens, and Z.~Wojna, ``Rethinking the
  inception architecture for computer vision,'' in \emph{Proceedings of the
  IEEE Conference on Computer Vision and Pattern Recognition}, 2016, pp.
  2818--2826.

\bibitem{howard2017mobilenets}
A.~G. Howard, M.~Zhu, B.~Chen, D.~Kalenichenko, W.~Wang, T.~Weyand,
  M.~Andreetto, and H.~Adam, ``Mobile{N}ets: Efficient convolutional neural
  networks for mobile vision applications,'' \emph{arXiv preprint
  arXiv:1704.04861}, 2017.

\bibitem{ioffe2015batch}
S.~Ioffe and C.~Szegedy, ``Batch normalization: Accelerating deep network
  training by reducing internal covariate shift,'' in \emph{International
  Conference on Machine Learning}, 2015, pp. 448--456.

\bibitem{golik2013cross}
P.~Golik, P.~Doetsch, and H.~Ney, ``Cross-entropy vs. squared error training: a
  theoretical and experimental comparison,'' in \emph{Interspeech}, 2013, pp.
  1756--1760.

\bibitem{levina2001earth}
E.~Levina and P.~Bickel, ``The earth mover's distance is the {M}allows'
  distance: Some insights from statistics,'' in \emph{Computer Vision, 2001.
  ICCV 2001. Proceedings. Eighth IEEE International Conference on},
  vol.~2.\hskip 1em plus 0.5em minus 0.4em\relax IEEE, 2001, pp. 251--256.

\bibitem{abadi2016tensorflow}
M.~Abadi, A.~Agarwal, P.~Barham, E.~Brevdo, Z.~Chen, C.~Citro, G.~S. Corrado,
  A.~Davis, J.~Dean, M.~Devin \emph{et~al.}, ``Tensor{F}low: Large-scale
  machine learning on heterogeneous distributed systems,'' \emph{arXiv preprint
  arXiv:1603.04467}, 2016.

\bibitem{abadi2016tensorflow2}
M.~Abadi, P.~Barham, J.~Chen, Z.~Chen, A.~Davis, J.~Dean, M.~Devin,
  S.~Ghemawat, G.~Irving, M.~Isard \emph{et~al.}, ``Tensor{F}low: A system for
  large-scale machine learning,'' in \emph{Proceedings of the 12th USENIX
  Symposium on Operating Systems Design and Implementation (OSDI). Savannah,
  Georgia, USA}, 2016.

\bibitem{xu2016blind}
J.~Xu, P.~Ye, Q.~Li, H.~Du, Y.~Liu, and D.~Doermann, ``Blind image quality
  assessment based on high order statistics aggregation,'' \emph{IEEE
  Transactions on Image Processing}, vol.~25, no.~9, pp. 4444--4457, 2016.

\bibitem{lu2014rapid}
X.~Lu, Z.~Lin, H.~Jin, J.~Yang, and J.~Z. Wang, ``Rapid: Rating pictorial
  aesthetics using deep learning,'' in \emph{Proceedings of the 22nd ACM
  international conference on Multimedia}.\hskip 1em plus 0.5em minus
  0.4em\relax ACM, 2014, pp. 457--466.

\bibitem{kao2016visual}
Y.~Kao, R.~He, and K.~Huang, ``Visual aesthetic quality assessment with
  multi-task deep learning,'' \emph{arXiv preprint arXiv:1604.04970}, 2016.

\bibitem{wang2016brain}
Z.~Wang, S.~Chang, F.~Dolcos, D.~Beck, D.~Liu, and T.~S. Huang,
  ``Brain-inspired deep networks for image aesthetics assessment,'' \emph{arXiv
  preprint arXiv:1601.04155}, 2016.

\bibitem{moorthy2011blind}
A.~K. Moorthy and A.~C. Bovik, ``Blind image quality assessment: From natural
  scene statistics to perceptual quality,'' \emph{IEEE transactions on Image
  Processing}, vol.~20, no.~12, pp. 3350--3364, 2011.

\bibitem{mittal2012no}
A.~Mittal, A.~K. Moorthy, and A.~C. Bovik, ``No-reference image quality
  assessment in the spatial domain,'' \emph{IEEE Transactions on Image
  Processing}, vol.~21, no.~12, pp. 4695--4708, 2012.

\bibitem{saad2012blind}
M.~A. Saad, A.~C. Bovik, and C.~Charrier, ``Blind image quality assessment: A
  natural scene statistics approach in the {DCT} domain,'' \emph{IEEE
  Transactions on Image Processing}, vol.~21, no.~8, pp. 3339--3352, 2012.

\bibitem{kottayil2016color}
N.~K. Kottayil, I.~Cheng, F.~Dufaux, and A.~Basu, ``A color intensity invariant
  low-level feature optimization framework for image quality assessment,''
  \emph{Signal, Image and Video Processing}, vol.~10, no.~6, pp. 1169--1176,
  2016.

\bibitem{talebi2016fast}
H.~Talebi and P.~Milanfar, ``Fast multilayer {L}aplacian enhancement,''
  \emph{IEEE Transactions on Computational Imaging}, vol.~2, no.~4, pp.
  496--509, 2016.

\bibitem{wong2016turbo}
T.-S. Wong and P.~Milanfar, ``Turbo denoising for mobile photographic
  applications,'' in \emph{Image Processing (ICIP), 2016 IEEE International
  Conference on}.\hskip 1em plus 0.5em minus 0.4em\relax IEEE, 2016, pp.
  988--992.

\bibitem{gastal2011domain}
E.~S. Gastal and M.~M. Oliveira, ``Domain transform for edge-aware image and
  video processing,'' in \emph{ACM Transactions on Graphics (ToG)}, vol.~30,
  no.~4.\hskip 1em plus 0.5em minus 0.4em\relax ACM, 2011, p.~69.

\bibitem{zhu2009no}
X.~Zhu and P.~Milanfar, ``A no-reference sharpness metric sensitive to blur and
  noise,'' in \emph{Quality of Multimedia Experience, 2009. QoMEx 2009.
  International Workshop on}.\hskip 1em plus 0.5em minus 0.4em\relax IEEE,
  2009, pp. 64--69.

\bibitem{zhu2010automatic}
------, ``Automatic parameter selection for denoising algorithms using a
  no-reference measure of image content,'' \emph{IEEE Transactions on Image
  Processing}, vol.~19, no.~12, pp. 3116--3132, 2010.

\end{thebibliography}

\end{document}